\def\year{2017}\relax
\documentclass[letterpaper]{article}
\usepackage{fullpage}
\usepackage{times}
\usepackage{helvet}
\usepackage{courier}
\usepackage{amssymb, amsfonts, amsmath}
\usepackage{graphicx, color}
\usepackage{array}
\usepackage{cite}
\usepackage{epstopdf}
\usepackage{calrsfs}

\DeclareMathAlphabet{\pazocal}{OMS}{zplm}{m}{n}

\newtheorem{lem}{Lemma}

\usepackage{tikz}
\usetikzlibrary{shapes, arrows, positioning, decorations.markings}

\definecolor {processblue}{cmyk}{0.96,0,0,0}
\tikzstyle{int}=[draw, fill=blue!20, minimum size=2em]
\tikzstyle{init} = [pin edge={to-,thin,black}]

\usetikzlibrary{shapes}
\usetikzlibrary{fit}
\usetikzlibrary{chains}
\usetikzlibrary{arrows}

\tikzstyle{plate} = [draw, rectangle, rounded corners, fit=#1]
\tikzstyle{wrap} = [inner sep=0pt, fit=#1]

\tikzstyle{caption} = [node distance=0] %
\tikzstyle{bottom plate caption} = [caption, node distance=0, inner sep=0pt,
below left=-5pt and 0pt of #1.south east] %

\tikzstyle{top plate caption} = [caption, node distance=0, inner sep=0pt,
below left=0pt and 0pt of #1.north east] %

\begin{document}

\title{Prototypical Recurrent Unit}
\date{}
\author{Dingkun Long${^1}$, Richong Zhang${^1}$, Yongyi Mao${^2}$\\
${^1}$School of Computer Science and Engineering, Beihang University, Beijing,China\\
${^2}$School of Electrical Engineering and Computer Science, University of Ottawa\\
\{longdk, zhangrc\}@act.buaa.edu.cn, yymao@eecs.uottawa.ca
}
\maketitle
\begin{abstract}

% The difficulty in analyzing LSTM-like recurrent neural networks lies in the complex structure of the recurrent unit, which induces highly complex nonlinear dynamics. In this paper, we design a new simple recurrent unit, which we call 
% Prototypical Recurrent Unit (PRU).  We verify experimentally that PRU performs comparably to LSTM and GRU. This potentially enables PRU to be a prototypical example for analytic study of LSTM-like recurrent networks.  Along these experiments, the memorization capability of LSTM-like networks is also studied and some insights are obtained.

Despite the great successes of deep learning, the effectiveness of deep neural networks has not been understood at any theoretical depth.  This work is motivated by the thrust of developing a deeper understanding of recurrent neural networks, particularly LSTM/GRU-like networks. As the highly complex structure of the recurrent unit in LSTM and GRU networks makes them difficult to analyze, our methodology in this research theme is to construct an alternative recurrent unit that is as simple as possible and yet also captures the key components of LSTM/GRU recurrent units. Such a unit can then be used for the study of recurrent networks and its structural simplicity may allow easier analysis.  Towards that goal, we take a system-theoretic perspective to design a new recurrent unit, which we call the prototypical recurrent unit (PRU).  Not only having minimal complexity, PRU is demonstrated experimentally to have comparable performance to GRU and LSTM unit. This establishes PRU networks as a prototype for future study of LSTM/GRU-like recurrent networks. This paper also studies the memorization abilities of LSTM, GRU and PRU networks, motivated by the folk belief that such networks possess long-term memory.  For this purpose, we design a simple and controllable task, called ``memorization problem'',  where the networks are trained to memorize certain targeted information. We show that the memorization performance of all three networks depends on the amount of targeted information, the amount of ``interfering" information, and the state space dimension of the recurrent unit. Experiments are also performed for another controllable task, the adding problem, and similar conclusions are obtained. 

\end{abstract}

%\noindent Congratulations on having a paper selected for inclusion in an AAAI Press proceedings or technical report! This document details the requirements necessary to get your accepted paper published using \LaTeX{}. If you are using Microsoft Word, instructions are provided in a different document. If you want to use some other formatting software, you must obtain permission from AAAI Press first.

%\vspace{-0.5cm}

\section{Introduction}

Deep learning has demonstrated great power in the recent years and appears to have prevailed in a broad spectrum of application domains (see, e.g., \cite{HintonSalakhutdinov_Science_2006:deepBeliefNet} \cite{lecun2015deepNatureReview}).  Despite its great successes, the effectiveness of deep neural networks has not been understood at a theoretical depth.  Thus developing novel analytic tools and theoretical frameworks for studying deep neural networks is of the greatest importance at the present time, and is anticipated to be a central subject of machine learning research in the years to come.

This work is motivated by the thrust of understanding recurrent neural networks, particularly LSTM/GRU-like networks \cite{hochreiter1997long} \cite{gers2000learning} \cite{peephole2002learning} \cite{cho2014learning} \cite{van2016pixel}. These networks have demonstrated to be the state-of-the-art models for time series or sequence data \cite{graves2013speech} \cite{bahdanau2014neural} \cite{serban2016building}. Recently LSTM/GRU recurrent units have also been successfully adopted for modelling other forms of data (e.g., \cite{cheng2016long} \cite{srivastava2015unsupervised}). Despite these successes, the design of LSTM and GRU recurrent units was in fact heuristical;  to date there is little theoretical analysis justifying their effectiveness.  A particularly interesting observation regarding these networks is that they appear to possess ``long-term memory'', namely, being able to selectively ``remember'' the information from many time steps ago \cite{gers2001long}. As one may naturally expect such memorization capability  to have played an important role in the working of these networks, this aspect has not been well studied, analytically or experimentally.

The difficulty in analyzing recurrent networks resides in the complex structure of the recurrent unit, which induces highly complex nonlinear dynamics.  To understand LSTM-like recurrent networks, the methodology explored in this theme of research is to maximally simplify the structure of the recurrent unit. That is,  we wish to
construct an alternative recurrent unit that captures the key components LSTM and GRU but stays as simple as possible. Such a unit can then be used for the study of recurrent networks and its structural simplicity may allow easier analysis in future research.

Towards that goal, the main objective of this present paper is to design such a recurrent unit and verify that this unit performs comparably to LSTM and GRU.  To that end, we develop a new recurrent unit, which we call the {\em Prototypical Recurrent Unit} (PRU). We rationalize our design methodology from a system-theoretic perspective where a recurrent unit is understood as a causal time-invariant system in state-space representations. Insights from previous research suggest that additive evolution appear essential for LSTM-like networks to avoid the ``gradient-vanishing'' problem under back-propagation \cite{jozefowicz2015empirical} \cite{chung2014empirical}  \cite{mikolov2014learning}. This understanding is also exploited in our design of PRU.

The performance of PRU is verified and compared against LSTM and GRU via extensive experiments. Using these three kinds of recurrent unit, we not only experiment on constructing a standard language model for character prediction \cite{mikolov2010recurrent}, but also test the recurrent units for two controlled learning tasks, the Adding Problem \cite{hochreiter1997long}, and the Memorization Problem. The latter problem is what we propose in this work specifically for studying the memorization capability of the recurrent networks.  All experimental results confirm that PRU performs comparably to LSTM and GRU, achieving the purpose of this paper.

As another contribution,  our experiments in this work demonstrate that the intrinsic memorization capability of the recurrent units depends critically on the dimension of the state space. The amount of targeted information (for memorization), the duration of memory, and the intensity of the interfering signal also directly impact the memorization performance.

Finally it is perhaps worth noting that although PRU is designed to be a prototype which hopefully allows for easier analysis in future research,  our experiments suggest that it can also be used as a practical alternative to LSTM and GRU. A particular advantage of PRU is its time complexity. In this metric, PRU demonstrates to be superior to both LSTM and GRU.

%\vspace{-0.2cm}

\section{State-Space Representations}

In system theory~\cite{Kha02}, a (discrete-time) system can be understood as any physical or conceptual device that responds to an {\em input sequence}  $x_1, x_2, \ldots $ and generates an {\em output sequence} $y_1, y_2, \ldots$, where the indices of the sequences are discrete time. In general, each $x_t$ and each $y_t$ at any time $t$ may be a vector of arbitrary dimensions. We will then use ${\pazocal X}$ and ${\pazocal Y}$ to denote the vector spaces from which $x_t$ and $y_t$ take value respectively. We will call ${\pazocal X}$ the {\em input space} and ${\pazocal Y}$ the {\em output space}.
The behaviour of the system is characterized by a function $J$ that maps the space of all input sequences to the space of all output sequences. Then two systems $J$ and $J'$ are {\em equivalent} if $J$ and $J'$ are identical as functions.

%For every causal system (namely, one in which the output $y_t$ at each time $t$ is independent of all future inputs $x_{t+1}, x_{t+2}, \ldots$),   the behaviour of the system can be described by a sequence of functions $J^{(1)}, J^{(2)}, \ldots$, where
%$J^{(n)}$ is a function mapping the $n$-fold cartesian product ${\pazocal X}^n$ of $\pazocal X$ to the $n$-fold cartesian product ${\pazocal Y}^n$ of $\pazocal Y$. That is, for each input sequence $(x_1, x_2, \ldots, x_n)$ up to time $n$, the output sequence $(y_1, y_2, \ldots, y_n)$ is given by
%\[
%(y_1, y_2, \ldots, y_n) = J^{(n)} (x_1, x_2, \ldots, x_n).
%\]
%A causal system can then be specified by the triple $({\pazocal X}, {\pazocal Y}, \{J^{(t)}:t\in {\mathbb N}\})$, where we have used $\mathbb N$ to denote the  set of all positive integers.  In this view,  two systems $A=({\pazocal X}_A, {\pazocal Y}_A, \{J_A^{(t)}:t\in {\mathbb N}\})$ and $B=({\pazocal X}_B, {\pazocal Y}_B, \{J_B^{(t)}:t\in {\mathbb N}\})$ are {\em equivalent} if and only if ${\pazocal X}_A={\pazocal X}_B$, ${\pazocal Y}_A={\pazocal Y}_B$ and $J_A^{(t)}=J_B^{(t)}$ for each $t\in {\mathbb N}$.

The class of systems that are of primary interest are causal systems, namely those in which the output $y_t$ at each time $t$ is independent of all future inputs $x_{t+1}, x_{t+2}, \ldots$. The grand idea in system theory is arguably the introduction of the notion of {\em state} to causal systems~\cite{Kha02}. This makes state-space models the central topic in system theory, resulting in wide and profound impact on system analysis and design. In a nutshell, the state configuration is an quantity internal to the system, serving as a complete summary of the all past inputs so that {\em given the current state, the current and future outputs are independent of all past inputs}.
%(see the top diagram of Figure \ref{fig:twoStateRepresentations}).

In this perspective, a recurrent unit can be regarded precisely as a causal time-invariant system in a state-space representation. We now formalize such a state-space representations.

At each time instant $t$, in addition to the input variable $x_t$ and output variable $y_t$, the representation of a recurrent unit also contains a state variable $s_t$, taking values in a vector space ${\pazocal S}$, which will be referred to as the {\em state space}.  Before the system is excited by the input, or at time $t=0$, it is assumed that the state variable $s_0$ takes certain initial configuration, which is assumed customarily to be the origin $0\in {\pazocal S}$.

\begin{figure}
\begin{center}
\begin{tabular}{c}
\scalebox{0.7}{
\begin{tikzpicture}[-latex ,auto ,node distance =1.3 cm and 2 cm, on grid ,
semithick ,
state/.style ={ circle ,top color =white , bottom color = red!20 ,
draw, red , text=red , minimum width =0.1cm},
box/.style ={rectangle ,top color =white , bottom color = processblue!20 ,
draw, processblue , text=blue , minimum width =1cm , minimum height = 0.8cm, rounded corners}]

\node[](s0){};
\node[box](s1)[right= of s0]{};
\node[box](s2)[right=of s1]{};
\node[box](s3)[right = of s2]{};
\node[box](s4)[right = of s3]{};
\node[](end)[right = of s4]{};
\path (s0) edge node {$s_0$}(s1); 
\path (s1) edge node {$s_1$}(s2);
\path (s2) edge node {$s_2$}(s3);
\path (s3) edge node {$s_3$}(s4);   
\path (s4) edge node {$s_4$}(end);   

\node[](y1)[above= of s1]{$y_1$};
\node[](y2)[above= of s2]{$y_2$};
\node[](y3)[above= of s3]{$y_3$};
\node[](y4)[above= of s4]{$y_4$};
\path (s1) edge (y1);
\path (s2) edge (y2);
\path (s3) edge (y3);   
\path (s4) edge (y4);   

\node[](x1)[below= of s1]{$x_1$};
\node[](x2)[below= of s2]{$x_2$};
\node[](x3)[below= of s3]{$x_3$};
\node[](x4)[below= of s4]{$x_4$};

\path (x1) edge (s1);
\path (x2) edge (s2);
\path (x3) edge (s3);
\path (x4) edge (s4);

%\node[](term_2)[left = of f2]{};
%\path (term_2) edge node {$x_2$} (f2); 

\end{tikzpicture}
}\\
\\
\scalebox{0.7}{
\begin{tikzpicture}[-latex ,auto ,node distance =1.3 cm and 2 cm, on grid ,
semithick ,
state/.style ={ circle ,top color =white , bottom color = red!20 ,
draw, red , text=red , minimum width =0.1cm},
box/.style ={rectangle ,top color =white , bottom color = processblue!20 ,
draw, processblue , text=blue , minimum width =0.1cm , minimum height = 0.1cm}]

\node[state](s0){$s_0$};
\node[state](s1)[right= of s0]{$s_1$};
\node[state](s2)[right=of s1]{$s_2$};
\node[state](s3)[right = of s2]{$s_3$};
\node[state](s4)[right = of s3]{$s_4$};
\node[](end)[right = of s4]{};
\path (s0) edge (s1); 
\path (s1) edge (s2);
\path (s2) edge (s3);
\path (s3) edge (s4);   
\path (s4) edge (end);   

\node[state](y1)[above= of s1]{$y_1$};
\node[state](y2)[above= of s2]{$y_2$};
\node[state](y3)[above= of s3]{$y_3$};
\node[state](y4)[above= of s4]{$y_4$};
\path (s1) edge (y1);
\path (s2) edge (y2);
\path (s3) edge (y3);   
\path (s4) edge (y4);   

\node[state](x1)[below= of s1]{$x_1$};
\node[state](x2)[below= of s2]{$x_2$};
\node[state](x3)[below= of s3]{$x_3$};
\node[state](x4)[below= of s4]{$x_4$};

\path (x1) edge (s1);
\path (x2) edge (s2);
\path (x3) edge (s3);
\path (x4) edge (s4);

\path (x1) edge [bend right =35] (y1);
\path (x2) edge [bend right =35] (y2);
\path (x3) edge [bend right =35] (y3);
\path (x4) edge [bend right =35] (y4);

%\node[](term_2)[left = of f2]{};
%\path (term_2) edge node {$x_2$} (f2); 

\end{tikzpicture}}\\
\\
\scalebox{0.7}{
\begin{tikzpicture}[-latex ,auto ,node distance =1.3 cm and 2 cm, on grid ,
semithick ,
state/.style ={ circle ,top color =white , bottom color = red!20 ,
draw, red , text=red , minimum width =0.1cm},
box/.style ={rectangle ,top color =white , bottom color = processblue!20 ,
draw, processblue , text=blue , minimum width =0.1cm , minimum height = 0.1cm}]

\node[state](s0){$s_0$};
\node[state](s1)[right= of s0]{$s_1$};
\node[state](s2)[right=of s1]{$s_2$};
\node[state](s3)[right = of s2]{$s_3$};
\node[state](s4)[right = of s3]{$s_4$};
\node[](end)[right = of s4]{};
\path (s0) edge (s1); 
\path (s1) edge (s2);
\path (s2) edge (s3);
\path (s3) edge (s4);   
\path (s4) edge (end);   

\node[state](y1)[above= of s1]{$y_1$};
\node[state](y2)[above= of s2]{$y_2$};
\node[state](y3)[above= of s3]{$y_3$};
\node[state](y4)[above= of s4]{$y_4$};
\path (s1) edge (y1);
\path (s2) edge (y2);
\path (s3) edge (y3);   
\path (s4) edge (y4);   

\node[state](x1)[below= of s1]{$x_1$};
\node[state](x2)[below= of s2]{$x_2$};
\node[state](x3)[below= of s3]{$x_3$};
\node[state](x4)[below= of s4]{$x_4$};

\path (x1) edge (s1);
\path (x2) edge (s2);
\path (x3) edge (s3);
\path (x4) edge (s4);

%\node[](term_2)[left = of f2]{};
%\path (term_2) edge node {$x_2$} (f2); 

\end{tikzpicture}
}
\end{tabular}
\end{center}
\vspace{-0.3cm}
\caption{\label{fig:twoStateRepresentations} A recurrent network(top) and the 
dependency structure of variables in Type-I representation (middle) and Type-II representation (bottom).}
\end{figure}
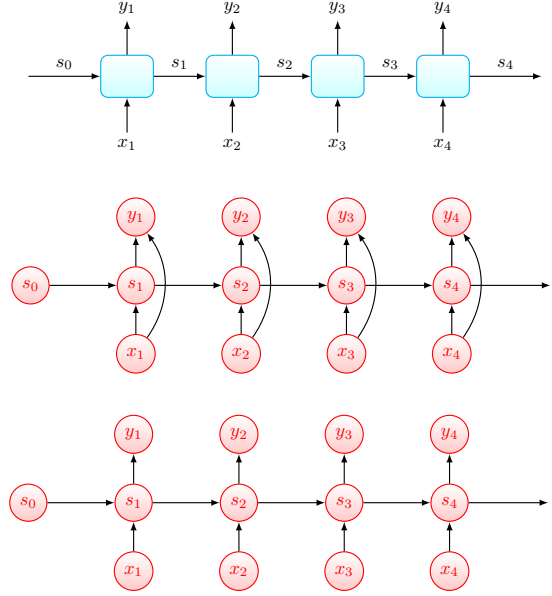

The behavior of the recurrent unit is governed by two
 functions $F:{\pazocal X}\times{\pazocal S} \rightarrow {\pazocal S}$ and $G:{\pazocal X}\times{\pazocal S} \rightarrow {\pazocal Y}$ as follows.  At each time instant $t$,
 function $F$ maps the current input $x_t$ and the previous state $s_{t-1}$ to the current state $s_t$, namely, via
\begin{equation}
\label{eq:general_F}
s_t=F(x_t, s_{t-1}),
\end{equation}
and function $G$ maps the current input $x_t$ and the current state $s_t$ to the current output $y_t$, namely,  via
\begin{equation}
\label{eq:general_G}
y_t=G(x_t, s_t).
\end{equation}

That is, in general a recurrent unit can be specified by the tuple $({\pazocal X}, {\pazocal Y}, {\pazocal S}, F, G)$ according to (\ref{eq:general_F}) and (\ref{eq:general_G}).  We call such specification of the recurrent unit  {\em Type-I state-space  representation} of the unit, and denote it by $({\pazocal X}, {\pazocal Y}, {\pazocal S}, F, G)_{\rm I}$.

It is remarkable that Type-I state-space representation is generic for any causal time-invariant system and hence generic for any recurrent unit. To illustrate this, we take the LSTM network as an example. The standard formulation of the LSTM network is given by the following equations:

\begin{eqnarray}
i_t & = &\sigma(W_i[c_{t-1},h_{t-1},x_t] + b_i)\\
f_t & = &\sigma(W_f[c_{t-1},h_{t-1},x_t] + b_f)\\
o_t & = &\sigma(W_o[c_{t-1},h_{t-1},x_t] + b_o)\\
\tilde{c}_t & = & tanh(W_c[h_{t-1},x_t] + b_g)\\
c_t & = & i_t \odot \tilde{c}_t + f_t \odot c_{t-1}\\
h_t & = & o_t \odot tanh(c_t)
\end{eqnarray}

where $\odot$ is the element-wise product. In these equations, if we take $(c_t, h_t)$ as state $s_t$, and $h_t$ as $y_t$, Equations(3-7) can be expressed as Equation (\ref{eq:general_F}), and Equations (5) and (8) can be expressed as Equation (\ref{eq:general_G}).  We then arrive at a Type-I representation.  It is also easy to verify that the recurrent unit in RNN\cite{elman1991distributed} and GRU networks can all be expressed this way.

As a clarification which might be necessary for the remainder of this paper, we pause to remark that in this paper (and under a system-theoretic perspective), the notion of a recurrent unit and that of a recurrent (neural) network are synonyms. In particular, a recurrent unit that operates over $n$ time instances may be viewed as $n$ copies of the same recurrent unit connected in a chain-structured network as shown in Figure \ref{fig:twoStateRepresentations} (top).   In this ``time-unfolded'' view,  the dependency structure between the variables in Type-I representation is shown in Figure \ref{fig:twoStateRepresentations} (middle).

Since we aim at designing a {\em simpler} recurrent unit, we now introduce another simpler representation, which we call {\em Type-II state-space representation}.  This representation is identical to the Type-I representation
except that the function $G$ is made to have domain ${\pazocal S}$, or alternatively put, the current output $y_t$ at each time $t$ is made dependent only of the current state $s_t$. That is,  $G$ acts only on $s_t$ and generates $y_t$ via
\begin{equation}
\label{eq:canonical_G}
y_t=G(s_t).
\end{equation}
Under this representation, the recurrent unit is specified again by the tuple $({\pazocal X}, {\pazocal Y}, {\pazocal S}, F, G)$, but according to (\ref{eq:general_F}) and (\ref{eq:canonical_G}). We denote this representation by $({\pazocal X}, {\pazocal Y}, {\pazocal S}, F, G)_{\rm II}$.  A diagram exhibiting the dependency structure of the variables in this representation is shown in Figure \ref{fig:twoStateRepresentations} (bottom).

The following lemma suggests that  Type-II representation has precisely the same expressive power as Type-I representation.

\begin{lem} Given its input and output spaces ${\pazocal X}$ and ${\pazocal Y}$,
 a recurrent unit can be represented by $({\pazocal X}, {\pazocal Y}, {\pazocal S}, F, G)_{\rm I}$  for some choice of
 ${\pazocal S}$, $F$ and $G$
  if any only if it can be represented by
$({\pazocal X}, {\pazocal Y}, \widetilde{\pazocal S}, f, g)_{\rm II}$ for some choice of $\widetilde{\pazocal S}$, $f$ and $g$.
\end{lem}

As the proof of this lemma contains certain insights into state-space representations, we sketch it here.

% We now sketch the proof of this lemma.

The ``if'' part of the proof is trivial, since function $G$ in Equation (\ref{eq:canonical_G}) is a special case of function $G$ in Equation (\ref{eq:general_G}). The ``only if'' part can be proved by construction, proceeded as follows. Let  $({\pazocal X}, {\pazocal Y}, {\pazocal S}, F, G)_{\rm I}$  be given. Define
$\widetilde{\pazocal S}:= {\pazocal X} \times {\pazocal S}$. Let function $f:{\pazocal X}\times \widetilde{\pazocal S}\rightarrow \widetilde{\pazocal S}$ be defined as follows: for each $(x, x', s) \in {\pazocal X}\times {\pazocal X} \times {\pazocal S} ={\pazocal X}\times \widetilde{\pazocal S}$, $f(x, x', s)=(x^*, s^*)\in {\pazocal X} \times {\pazocal S} =\widetilde{\pazocal S}$, where $x^*=0\in {\pazocal X}$ and $s^*=F(x, s)\in {\pazocal S}$. Define function $g:\widetilde{S}\rightarrow {\pazocal Y}$ as follows: for each $(x, s)\in {\pazocal X}\times {\pazocal S} = \widetilde{\pazocal S}$,
$g(x, s)=G(x, s)$. Now the lemma can be proved by identifying that systems $({\pazocal X}, {\pazocal Y}, {\pazocal S}, F, G)_{\rm I}$ and $({\pazocal X}, {\pazocal Y}, \widetilde{\pazocal S}, f, g)_{\rm II}$ are equivalent.   This latter fact can be easily established using proof by induction.

The significance of this lemma is that every recurrent unit can be represented using Type-II representation, in which the current output is made only dependent of the current state. In the proof of this result, we see that to convert a Type-I representation to a Type-II representation, it may require {\em increasing the dimension of the state space}. In the worst case, although often unnecessary in practice, one can make the state space $\widetilde{\pazocal S}$ equal to
the cartesian product ${\pazocal X}\times {\pazocal S}$ of the input space $\pazocal X$ and the state space
${\pazocal S}$  in the Type-I representation.

%\vspace{-0.2cm}

\section{Prototypical Recurrent Unit (PRU)}

Given that there is no loss of expressive power in Type-II representation,  to arrive at a simplified recurrent unit, we will stay within this representation. That is, for some given choices of vector spaces ${\pazocal X}$, ${\pazocal Y}$, and ${\pazocal S}$, we will design two functions $F:{\pazocal X}\times {\pazocal S} \rightarrow {\pazocal S}$ and $G:{\pazocal S} \rightarrow {\pazocal S}$ for $({\pazocal X}, {\pazocal Y}, {\pazocal S}, F, G)_{\rm II}$. It is our hope that the designed recurrent unit captures the essence of recurrent unit in LSTM and GRU networks, but stays as simple as possible.

From the previous literature \cite{pascanu2013difficulty}, the following properties of LSTM and GRU appear crucial for their effectiveness.
\begin{enumerate}
\item The recurrent unit behaves according to a nonlinear system, where the nonlinearity is induced by the use of nonlinear activation functions such as sigmoid, tanh, or ReLU functions.
\item The evolution from state $s_t$ to state $s_{t+1}$ is additive. It has been understood that such a property is critical for eliminating the problem of vanishing or blowing-up gradient in backpropagation.
\end{enumerate}

Based on this understanding, our design philosophy is to impose these two properties {\em minimally} on the recurrent unit.  Our hypothesis is that if these two properties are indeed essential, the resulting recurrent unit will behave in a way similar to GRU and LSTM recurrent units and can be used as a prototypical example for in-depth understanding of LSTM/GRU-like recurrent networks.

Such a design philosophy naturally results in the following new recurrent unit, which we call the Prototypical Recurrent Unit (PRU) and now describe.

We begin with some notations. We consider ${\pazocal X} = {\mathbb R}^m$, ${\pazocal Y}= {\mathbb R}^l$ and ${\pazocal S}={\mathbb R}^k$; all vectors are taken as column vectors; the sigmoid (logistic or soft-max) function will be denoted by $\sigma$; when an activation function ($\sigma$, ${\rm tanh}$, or any other function $h:{\mathbb R}\rightarrow {\mathbb R}$) applies to a vector, it acts on the vector element-wise and outputs a vector of the same length.

With these notations, we describe the functions $F$ and $G$ in $({\pazocal X}, {\pazocal Y}, {\pazocal S}, F, G)_{\rm II}$ that defines PRU.

\noindent \underline{\bf Function $F$:} The function $F$ is defined by the following sequence of function compositions involving two other variables $u_t\in {\pazocal S}$ and $c_t\in {\mathbb R}^k$ (we note that although here $c_t$ is a $k$-dimensional vector, it should not be interpreted as a state configuration in ${\pazocal S}$ due to its physical meaning).
\begin{equation}
u_t= {\rm tanh}\left(
U_{\rm s}s_{t-1} + U_{\rm x}x_t + b_{\rm u}
\right)
\end{equation}
where $U_{\rm s}$ is a $k\times k$ matrix, $U_{\rm x}$ is a $k\times m$ matrix, and $b_{\rm u}$ is a $k$-dimensional vector.
\begin{equation}
c_t= \sigma \left(
C_{\rm s}^Ts_{t-1} + C_{\rm x}^Tx_t + b_{\rm c}
\right)
\end{equation}
where $C_{\rm s}$ is a $k\times k$ matrix, $C_{\rm x}$ is a $k\times m$ matrix, and $b_{\rm c}$ is a $k$-dimensional vector.
\begin{equation}
s_t= c_t \odot s_{t-1} + (1-c_t) \odot u_t
\end{equation}
where $\odot$ is the element-wise product.

\noindent \underline{\bf Function $G$:} The function $G$ is defined as follows.
\begin{equation}
y_t=h(Ws_t+b)
\end{equation}
where $W$ is an $l\times k$ matrix, $b$ is a length $l$ vector, and $h$ is an activation function.  Depending on the applications and the physical meaning of output $y_t$, $h$ can be chosen as $\sigma$, ${\rm tanh}$, ReLU, or even the identity function.

%{\bf HEEERRERERE}
%
%
%
%inside each unit, we create an internal  {\em control} variable $c_t$, a vector of dimension $k$, in which each element takes values in the interval $(0, 1)$ and governs the convex combination in property \ref{enum:convex}.
%
%
%The control variable $c_t$ is defined as
%\begin{equation}
%c_t= \sigma \left(
%C_{\rm s}^Ts_{t-1} + C_{\rm x}^Tx_t + b_{\rm c},
%\right)
%\end{equation}
%where $C_{\rm s}$ is a $k\times k$ matrix, $C_{\rm x}$ is a $k\times m$ matrix, and $b_{\rm c}$ is a $k$-dimensional vector.
%
%
%The synthesized information from $s_{t-1}$ and $x_t$ is a vector of length $k$, denoted by $u_t$. That is, $u_t$ takes values from the same space as the state variables.  Let $u_t$ be defined as
%\begin{equation}
%u_t= {\rm tanh}\left(
%U_{\rm s}s_{t-1} + U_{\rm x}x_t + b_{\rm u}
%\right)
%\end{equation}
%where $U_{\rm s}$ is a $k\times k$ matrix, $U_{\rm x}$ is a $k\times m$ matrix, and $b_{\rm u}$ is a $k$-dimensional vector.
%
%The state variable $s_t$ is then defined as
%\begin{equation}
%s_t= c_t \odot s_{t-1} + (1-c_t) \odot u_t,
%\end{equation}
%where $\odot$ is the element-wise product.
%
%Finally, the output variable $y_t$ is defined as
%\begin{equation}
%y_t=h(Ws_t+b)
%\end{equation}
%where $W$ is an $l\times k$ matrix, $b$ is a length $l$ vector, and $h$ is an activation function acting on its input vector component-wise. Depending on the targeted output, $h$ can be chosen as $\sigma$, ${\rm tanh}$, ReLU, or even the identity function.
%
At this point, we have completely defined PRU, which is parameterized by $\theta:=\left(C_{\rm x}, C_{\rm s}, b_{\rm c}, U_{\rm x}, U_{\rm s}, b_{\rm u}, W, b\right)$.

%\vspace{-0.2cm}

\section{Experimental Study}

Our experimental study serves two purposes.

First, we wish to verify that the designed PRU behaves similarly as LSTM and GRU.  For this purpose,
experiments need to be performed not only for real-world applications, in which one has no control over the datasets, but also for certain meaningful tasks where we have full control over the data.  Such controllable tasks will allow a comparison of these recurrent units over arbitrary ranges of data parameter settings, so as to fully demonstrate the performances of the compared recurrent units and reduce the risk of being biased by the statistics of a particular dataset.

Second, we wish to take the opportunity to investigate a fundamental aspect of recurrent networks, namely, their memorization capabilities. It has been experimentally observed and intuitively justified that LSTM/GRU-like recurrent unit has ``long-term memory'' \cite{greff2015lstm}.  Motivate by such observations, we are interested in thoroughly studying the memorization capability of these recurrent units and understand what factors may influence their memorization performance.

As such, we consider four different learning tasks, where the recurrent networks are trained to solve four different problems: the Memorization Problem, the Adding Problem, the Character Prediction Problem, and the MNIST Image Classification Problem.  The Character Prediction Problem and the MNIST Image Classification Problem are two well-known problems in the real-world application domain \cite{mikolov2010recurrent} \cite{kim2015character}. The Adding Problem is a controllable task, first introduced in \cite{hochreiter1997long}. The Memorization Problem is also a controllable task that we introduce in this work, inspired by the idea of a similar task presented  in \cite{bengio1994learning}.

All models in these experiments have the architecture shown in the top diagram of Figure \ref{fig:twoStateRepresentations} with single layer. In Memorization problem and Adding problem, we use single layer structure, in the other two experiments we will use a two-layer stacked structure. In the description of the experiments, when we speak of ``state space dimension'', for both PRU and GRU, it refers to the length of the vector passed between two consecutive recurrent units in the diagram.  In LSTM networks, there are two vectors of the same length passed between two consecutive recurrent units. Although from a system-theoretic perspective, two times this length should be regarded as the state space dimension, this choice would put LSTM in disadvantage.  This is because the output of the unit depends only on one of the vectors. For this reason, for LSTM networks, the term ``state space dimension'' refers to half of the true state-space dimension.

Experiments on Memorization Problem and Adding Problem are performed on the computer(Intel(R) Core(TM) i5-4570 CPU @3.20Hz), whereas experiment on Character Prediction Problem and MNIST Image Classification Problem are performed on a GeForce GTX 970 GPU. Time cost is evaluated in unit of second.

\subsection{Memorization Problem}

To describe this problem, let us first imagine a ``memorization machine'' ${\cal M}_{\rm mem}$ that behaves as follows. For any given non-negative integers $I$ and $N$, an input sequence $x_1, x_2, \ldots, x_{I+N}$ of scalar values are fed to the machine, where for $t=1, 2, \ldots, I$,
$x_t$ takes on value in $\{+1, -1\}$ each with probability $1/2$, and for $t=I+1, I+2, \ldots, I+N$,
$x_t$ is drawn independently from a Gaussian distribution with zero mean and variance $\delta^2$.  After processing the input sequence, the machine generates an output vector $(x_1, x_2, \ldots, x_I)^T$of dimension $I$. That is, as a function, the machine ${\cal M}_{\rm mem}$ behaves according to
\[
{\cal M}_{\rm mem}(x_1, x_2, \ldots, x_{I+N})=(x_1, x_2, \ldots, x_I)^T.
\]

Then in the Memorization Problem, the objective is to  train a model that simulates the behaviour of   ${\cal M}_{\rm mem}$, namely,  capable of ``memorizing'' the ``$I$ bit'' ``targeted information'' in the beginning of the input sequence,  after $N$ symbols of ``noise'' or ``interfering signal'' enter the model.   Obviously, the Memorization Problem is configured by three parameters:
 $I$, $N$, and $\delta$, where $I$ represents the amount of targeted information, $N$ represents the duration of memory, and
 $\delta$ represents the intensity of noise that  might interfere with the memorization behaviour of the model.

\noindent \underline{\bf Modelling:} Under a recurrent network model, it is natural to regard the input space ${\pazocal X}$ as ${\mathbb R}$ and the output space  ${\pazocal Y}$ as ${\mathbb R}^I$, and one may freely configure the dimension $k$ of the state space ${\pazocal S}$.  Except at the final time $t=I+N$, the output $y_t$ is discarded, and final output $y_{I+N}$ is used to simulate the output ${\cal M}_{\rm mem}(x_1, x_2, \ldots, x_{I+N})$ of the memorization machine.

\noindent \underline{\bf Datasets:} For each problem setting $(I, N, \delta^2)$, we generate 50000 training examples and 1000 testing examples according to the specification of the problem.

\noindent \underline{\bf Training:}  The training of each model is performed by optimizing the Mean Square Error (MSE) defined as
%The $y_t$ has dimension $I$, but the outputs of the units are not used except for the final unit, namely,  the one at $t=I+N$.  The output $Y_{I+N}$ has targets at generating vector $[X_1, X_2, \ldots, X_I]^T$.
\begin{equation}
\label{eq:MSE}
{\cal E}_{\rm MSE}(\theta): ={\mathbb E} \Vert{\cal M}_{\rm mem}(x_1, x_2, \ldots, x_{I+N}) - y_{I+N}\Vert^2
%\left[\left\Vert Y_{I+N} - [X_1, X_2, \ldots, X_I]^T\right\Vert^2\right]
\end{equation}
where the expectation operation $\mathbb E$ is taken as averaging over the training examples.  Mini-batched Stochastic Gradient Descent (SGD, in fact more precisely, mini-batched Back-Propagation Through Time) is used for this optimization. The batch size is chosen as $100$, the learning rate as $10^{-3}$, and the number of epochs as $1000$. Each component of the model parameters is initialized to random values drawn independently from the zero-mean unit-variance Gaussian distribution.

\noindent \underline{\bf Evaluation Metrics:} A trained model is evaluated using MSE defined in (\ref{eq:MSE}), where the expectation operation $\mathbb E$ is taken as averaging over the testing examples. For experiment setting $(I, N, \delta^2, k)$,  each studied model is trained $50$ times with different random initializations, and the average MSE is taken the performance metric for the experiment setting.  Time complexity for the three models are also evaluated.

\begin{figure*}[htb!]\centering
\begin{tabular}{ccc}
\includegraphics[width=.30\textwidth]{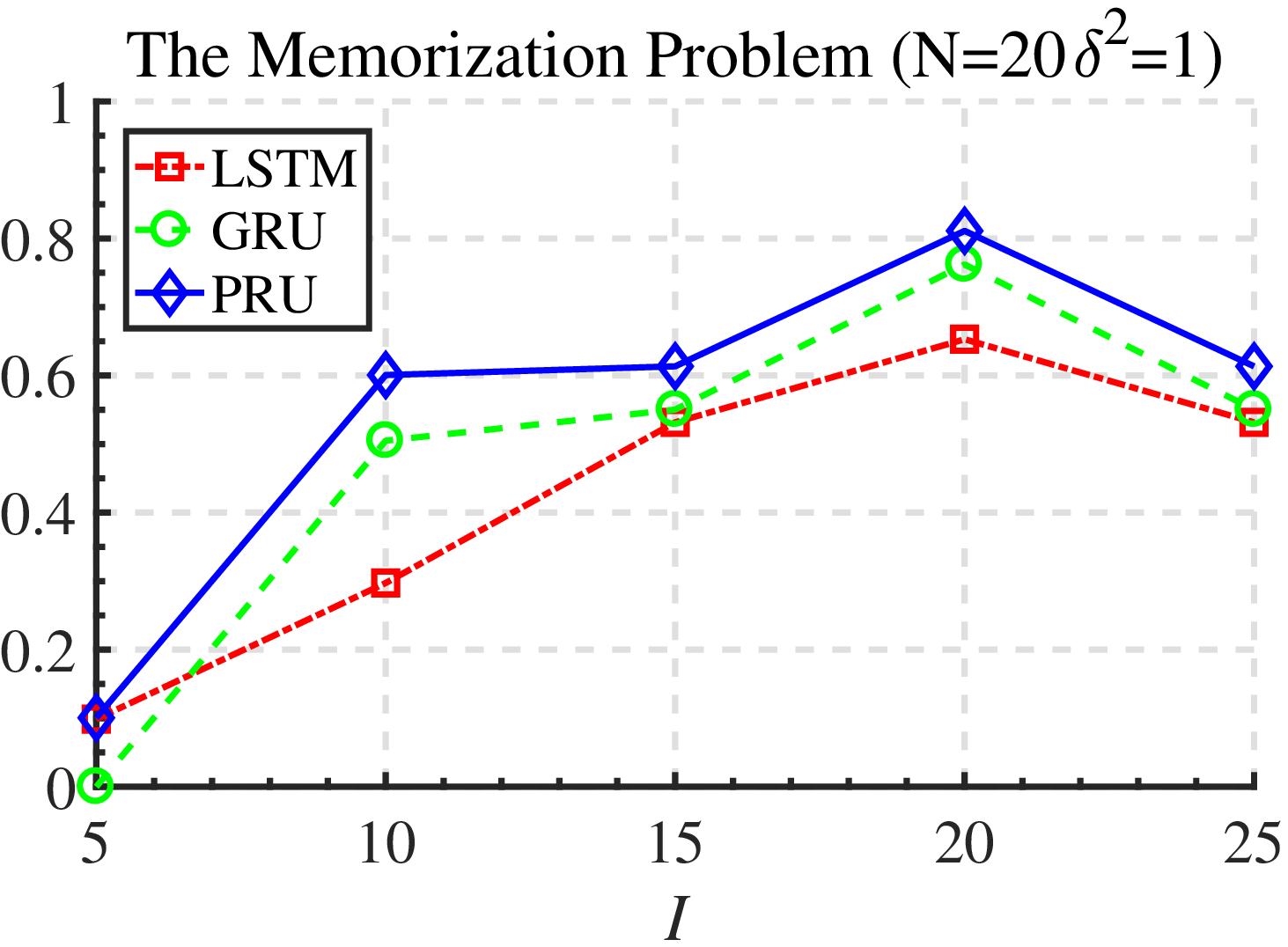} &
\includegraphics[width=.30\textwidth]{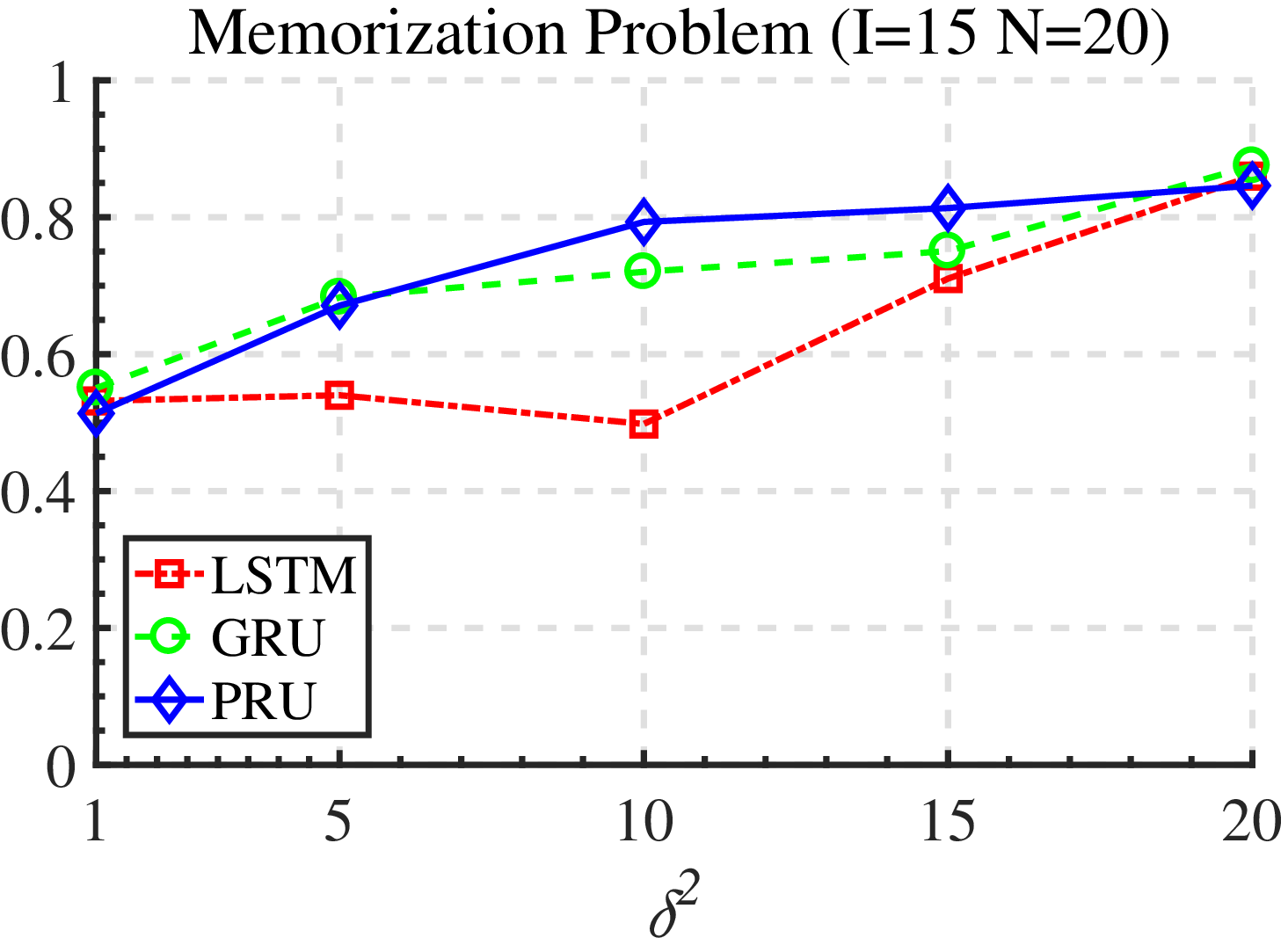} &
\includegraphics[width=.30\textwidth]{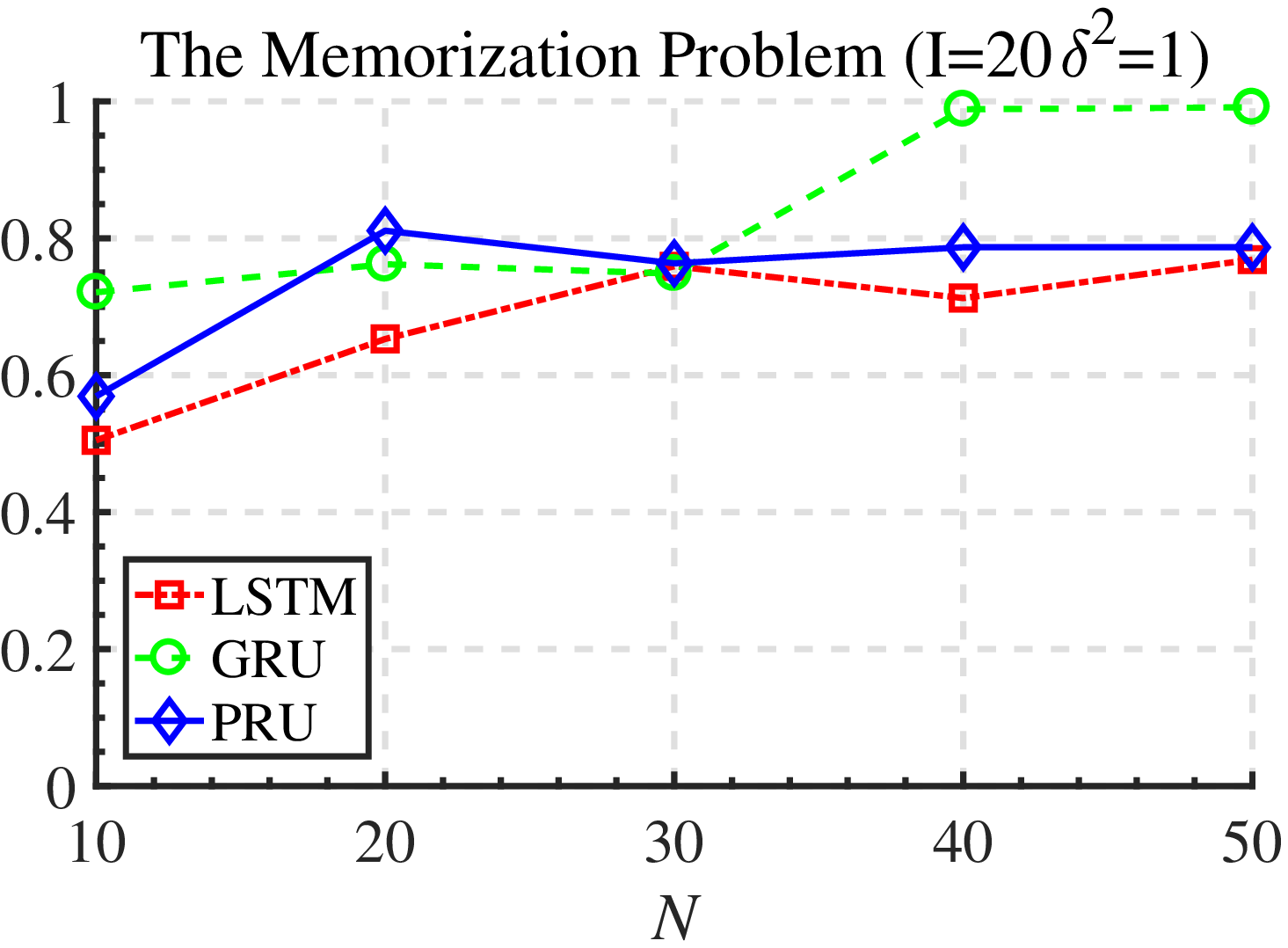}\\
\end{tabular}
\caption{MSE comparison of LSTM, GRU and PRU networks in Memorization Problem with same state space dimension}
\label{fig:overall_MP}
\end{figure*}

\begin{figure*}[htb!]\centering
\begin{tabular}{ccc}
\includegraphics[width=.30\textwidth]{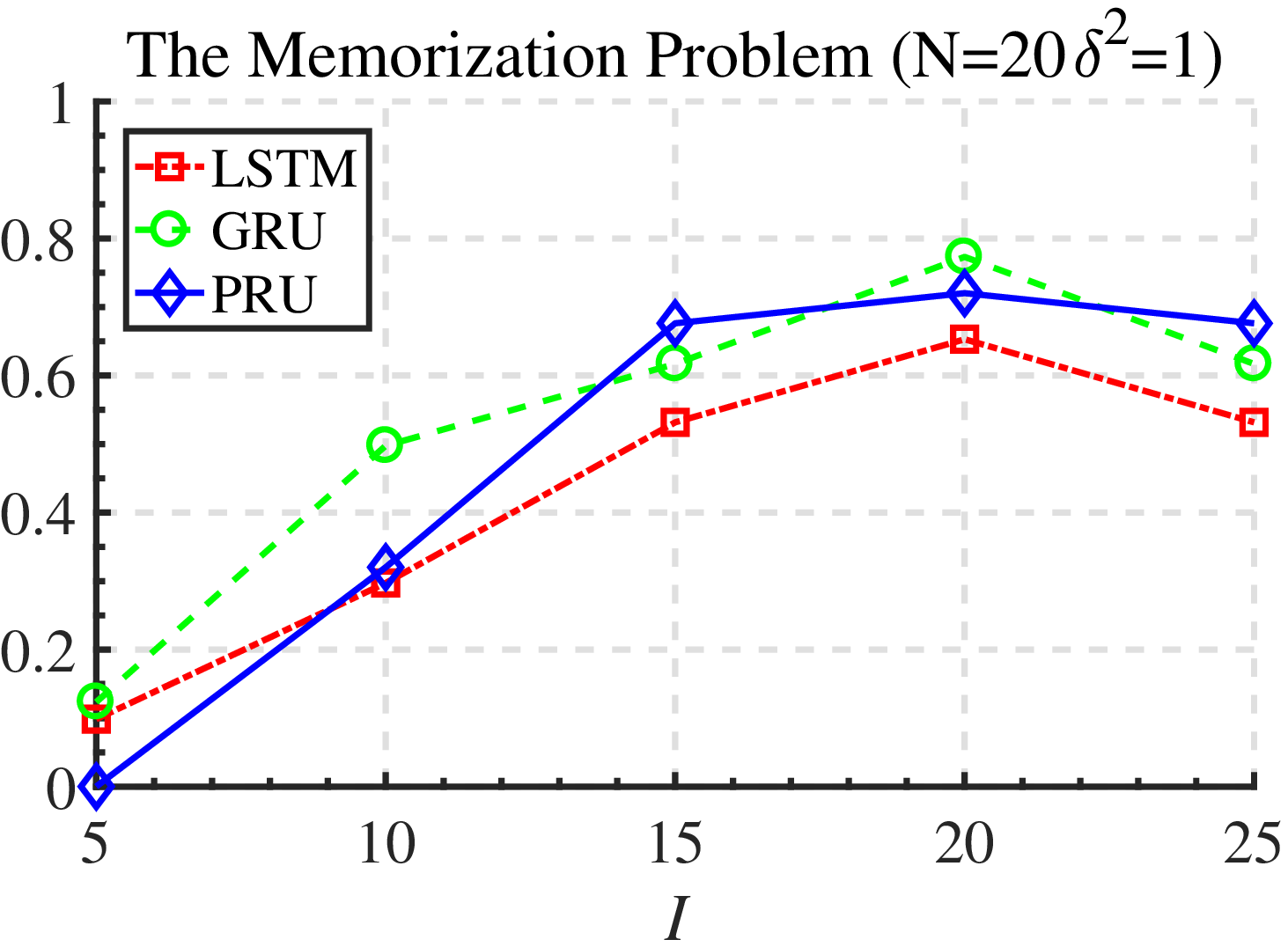} &
\includegraphics[width=.30\textwidth]{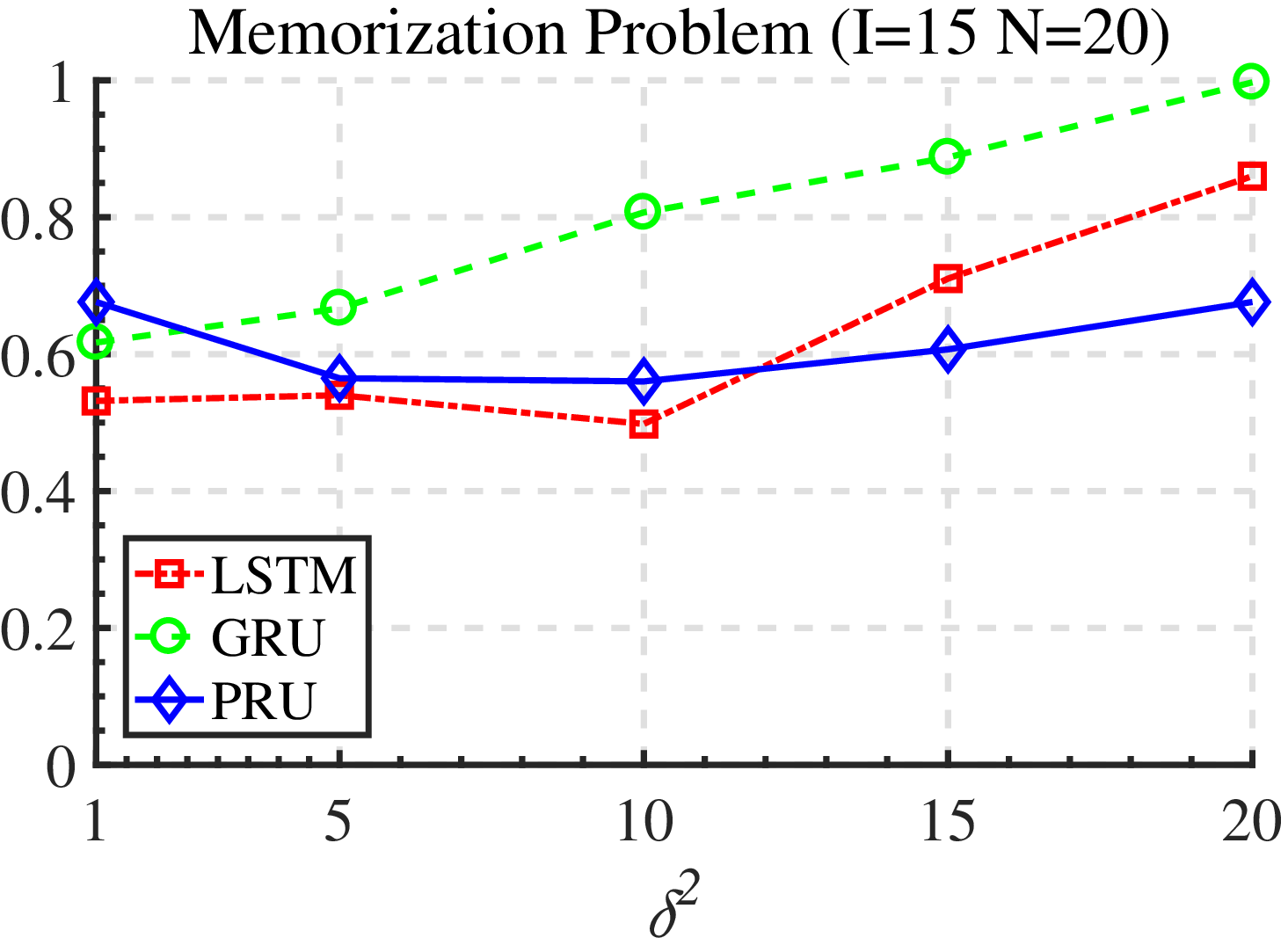} &
\includegraphics[width=.30\textwidth]{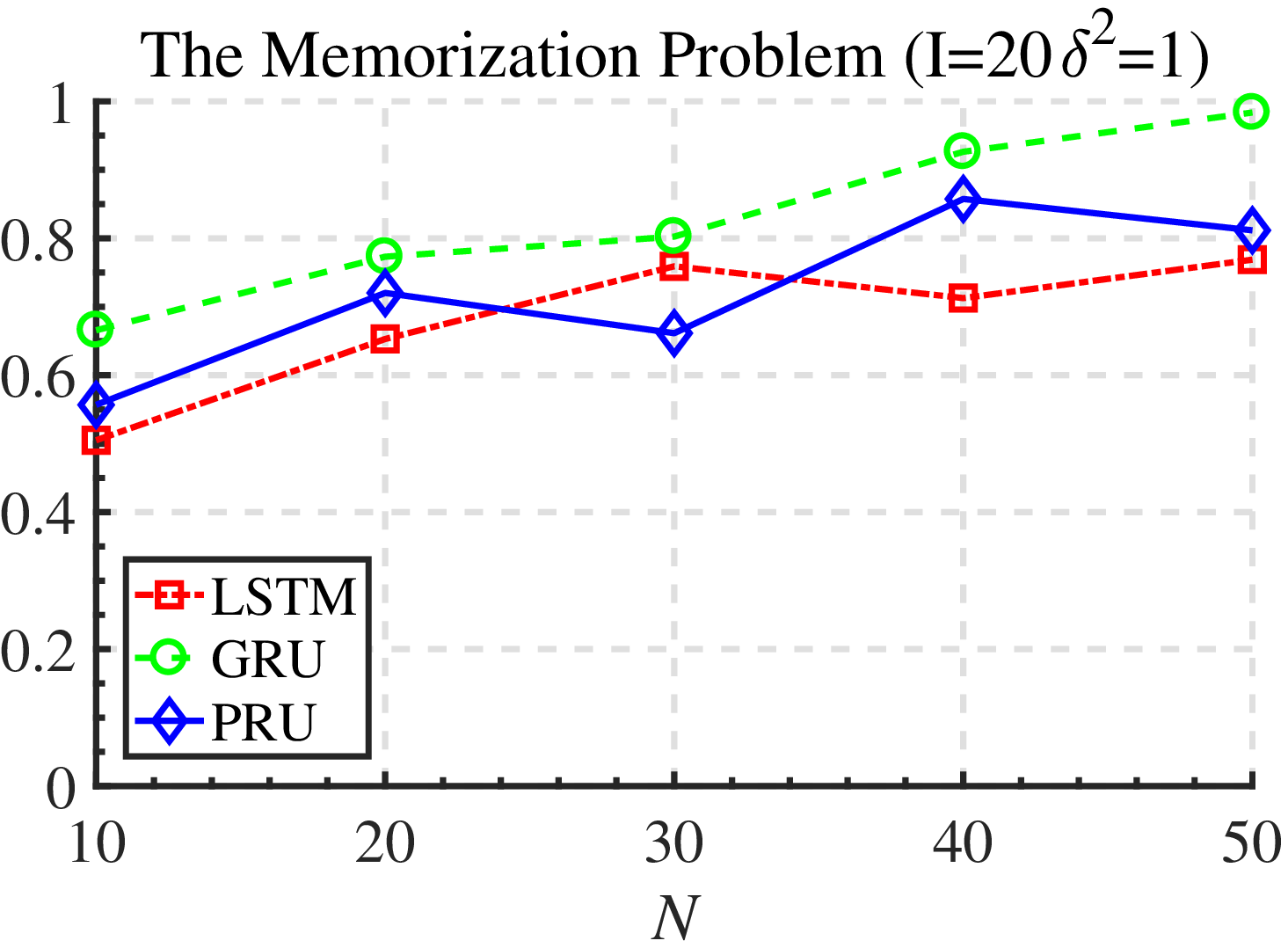}
\end{tabular}
\caption{\label{fig:mpsame parameter} MSE comparison of LSTM, GRU and PRU networks in Memorization Problem with same number of parameters.}
\end{figure*}

\begin{figure*}[htb!]\centering
\begin{tabular}{ccc}
\includegraphics[width=.30\textwidth]{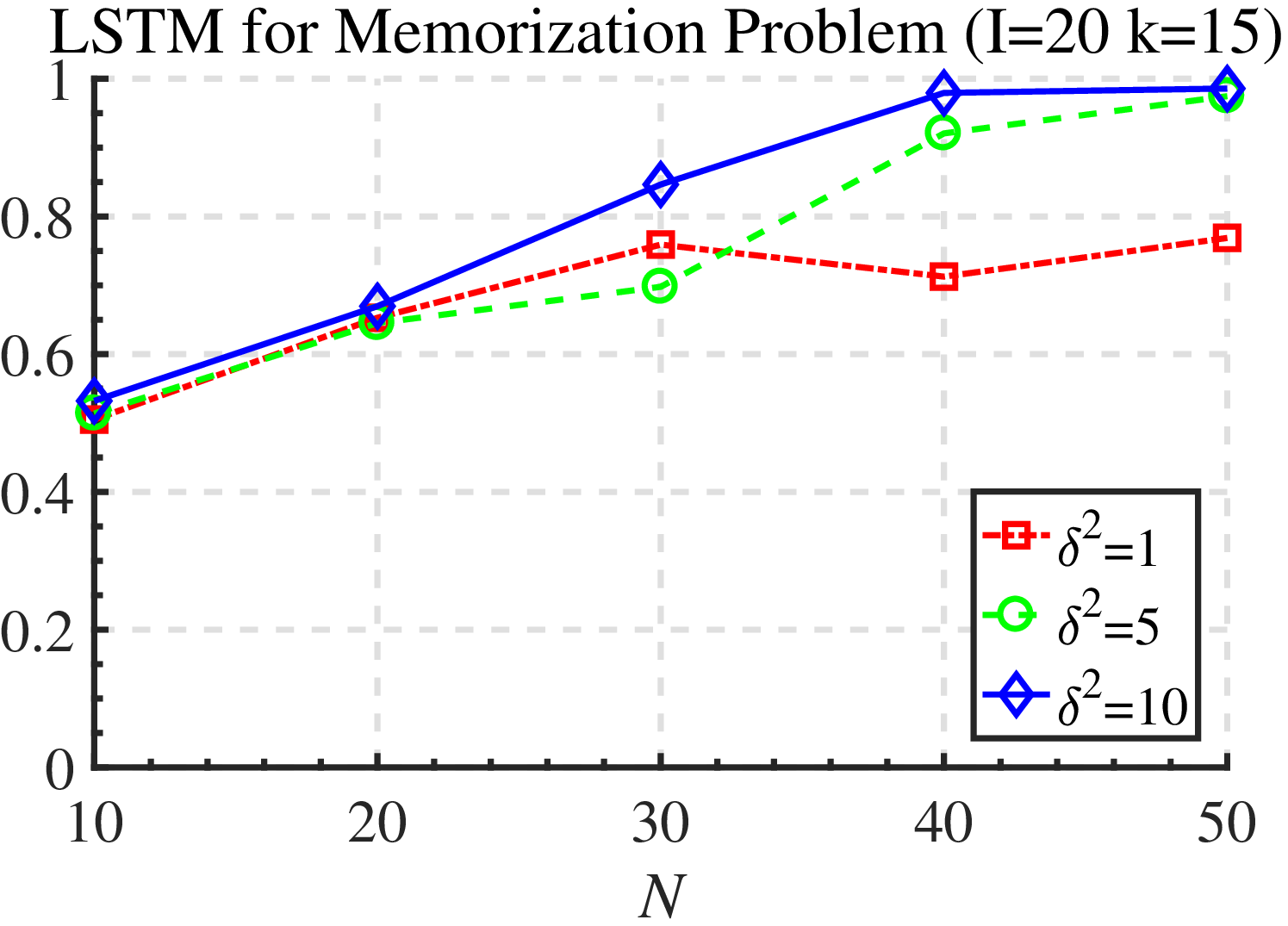} &
\includegraphics[width=.30\textwidth]{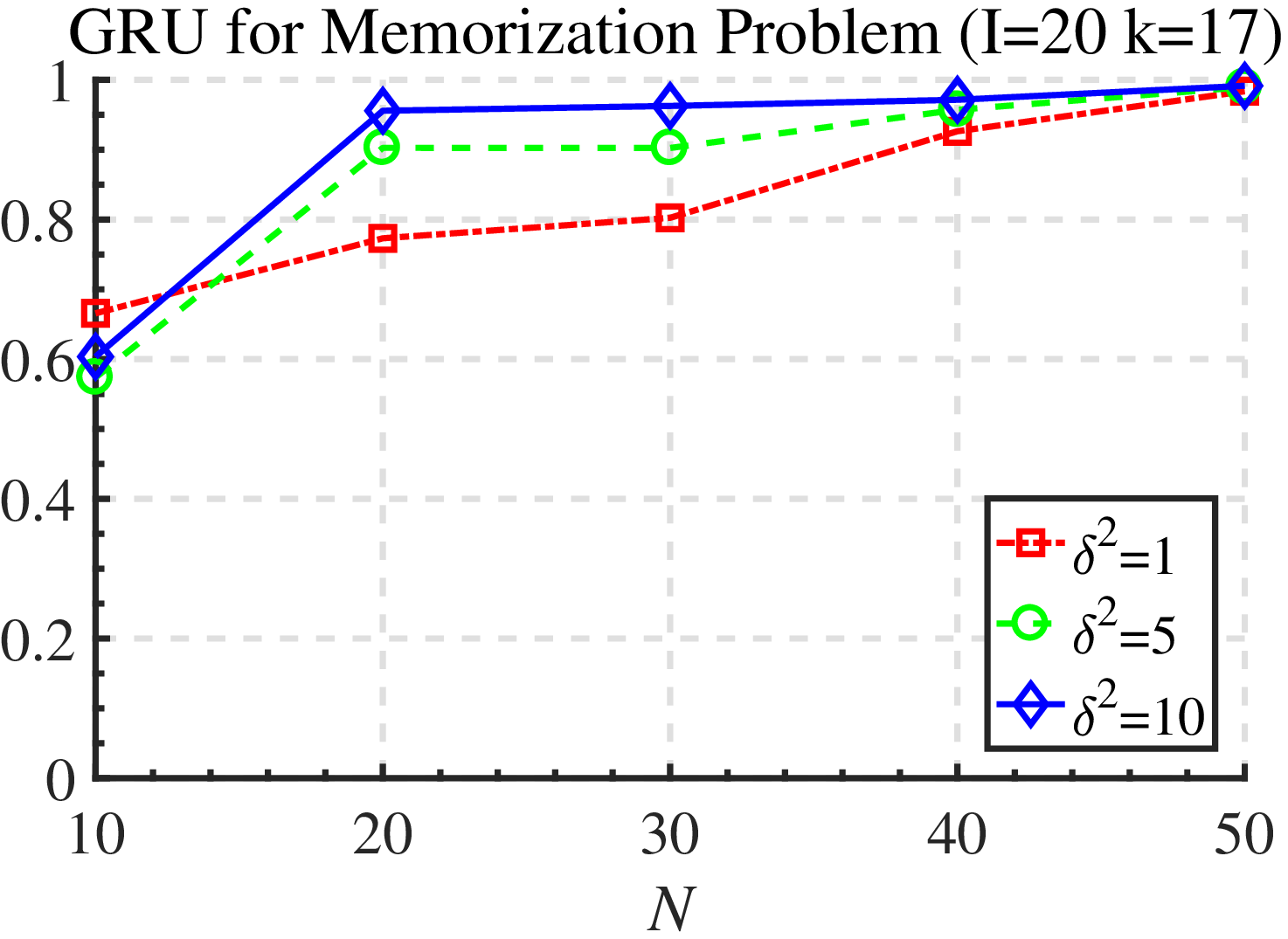} &
\includegraphics[width=.30\textwidth]{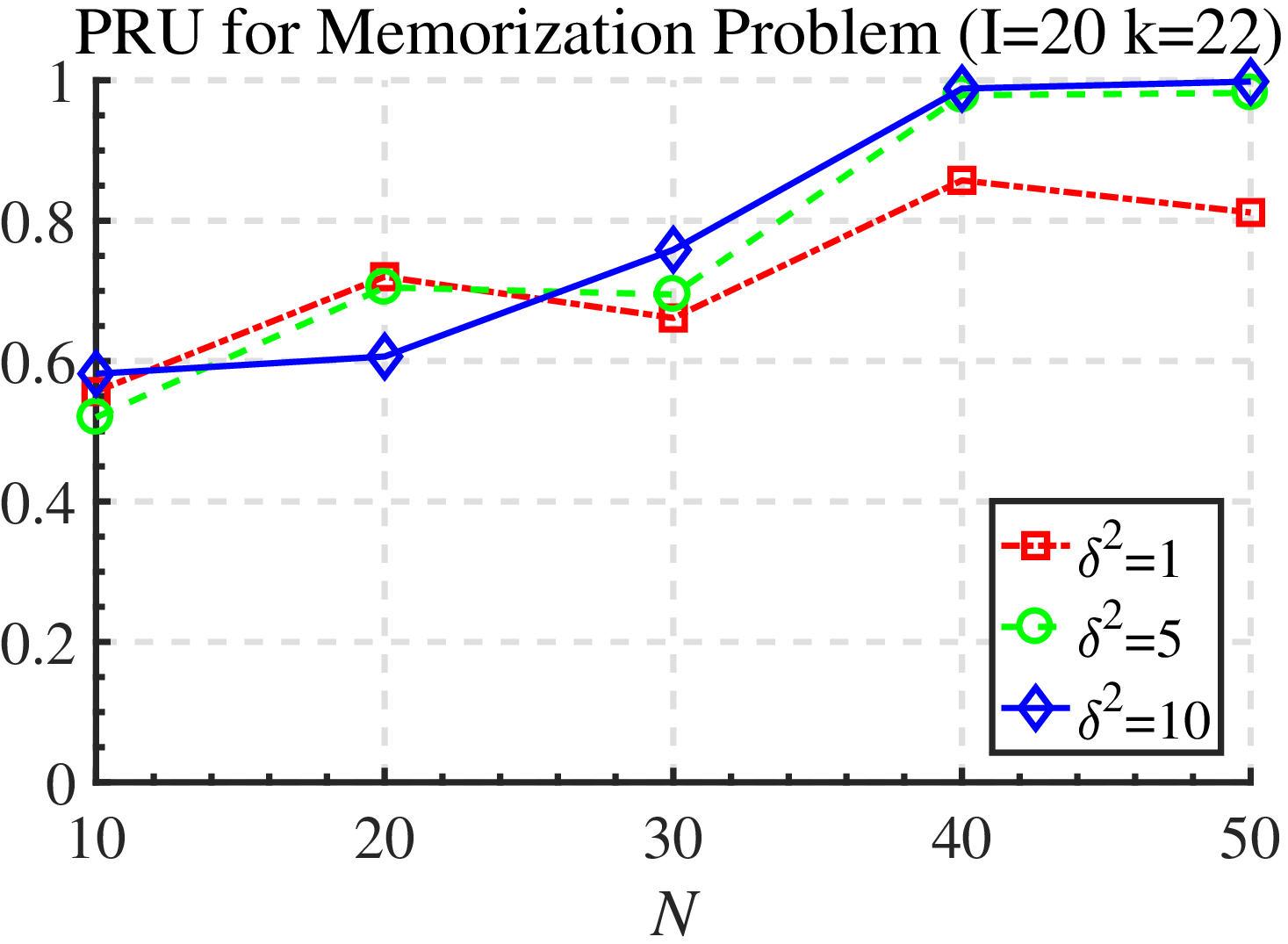}\\
\includegraphics[width=.30\textwidth]{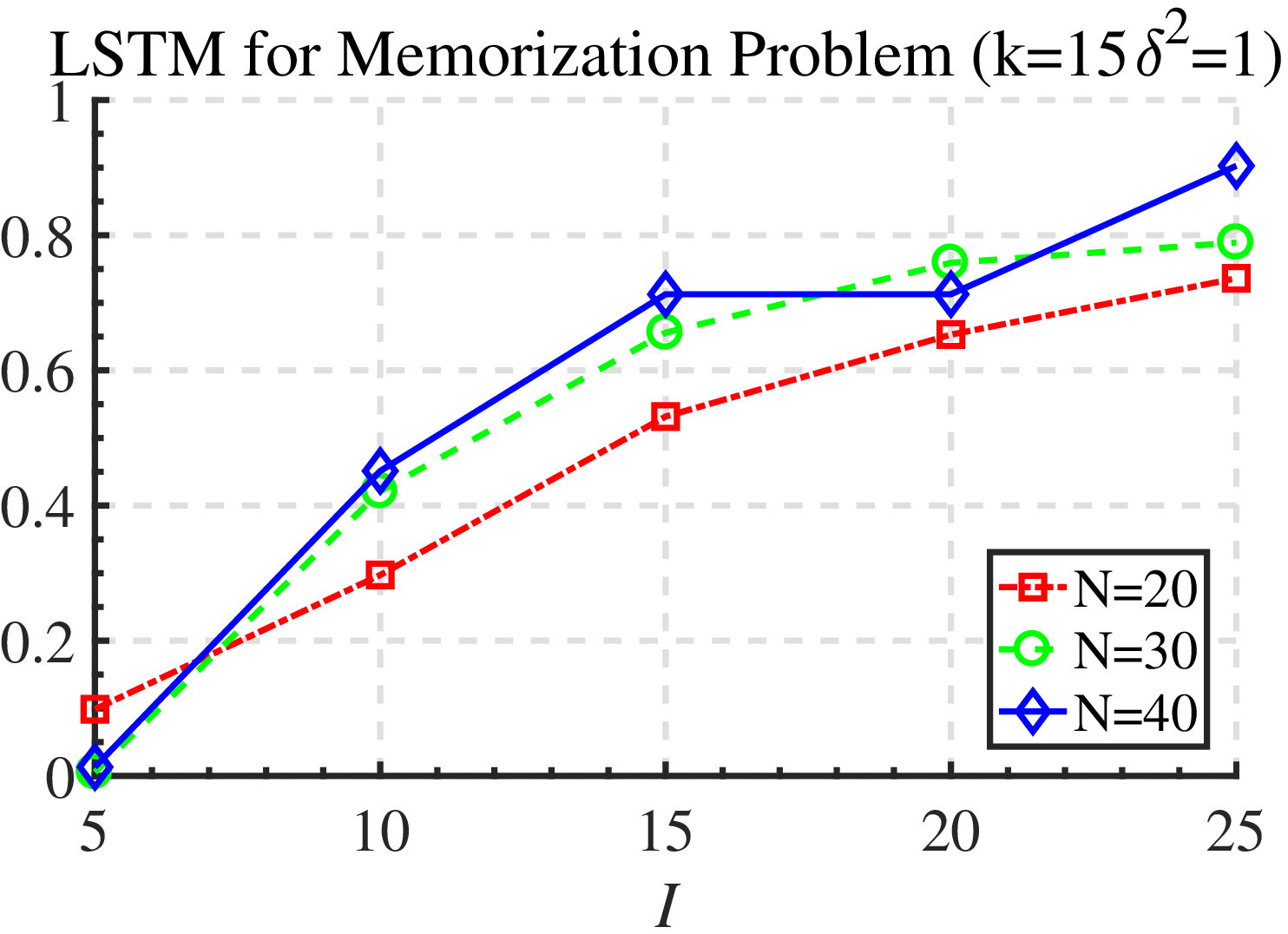} &
\includegraphics[width=.30\textwidth]{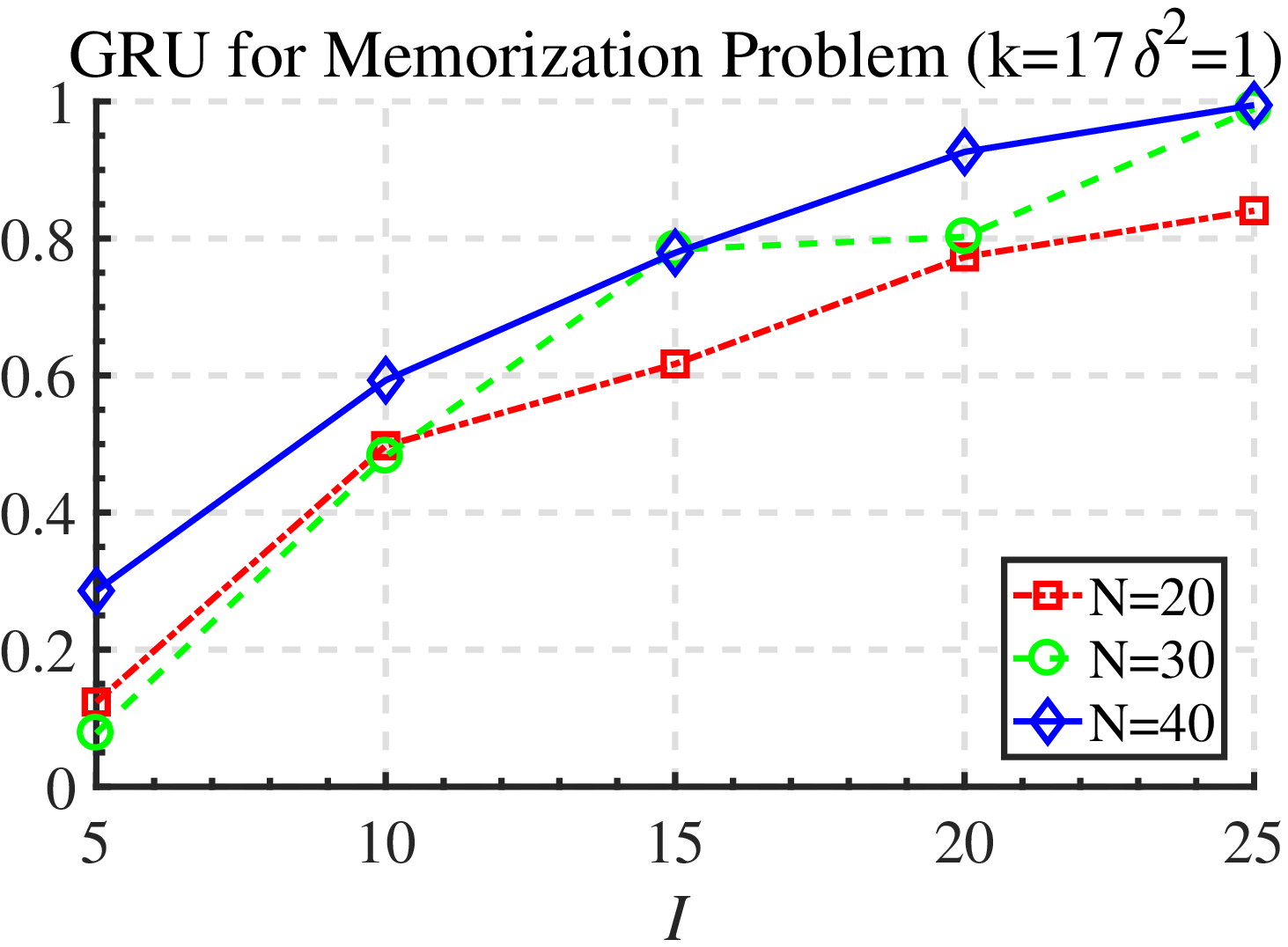} &
\includegraphics[width=.30\textwidth]{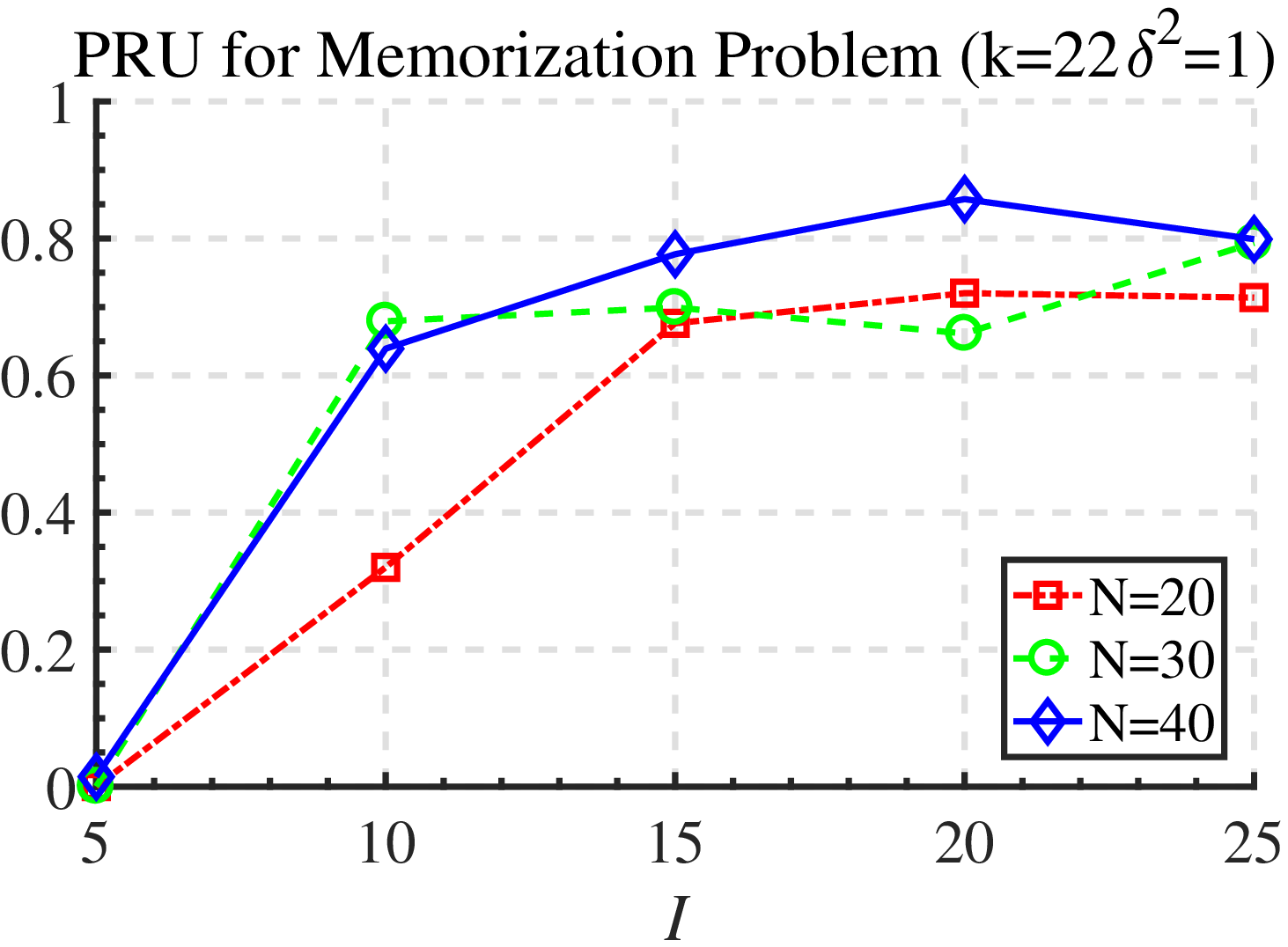}\\
\includegraphics[width=.30\textwidth]{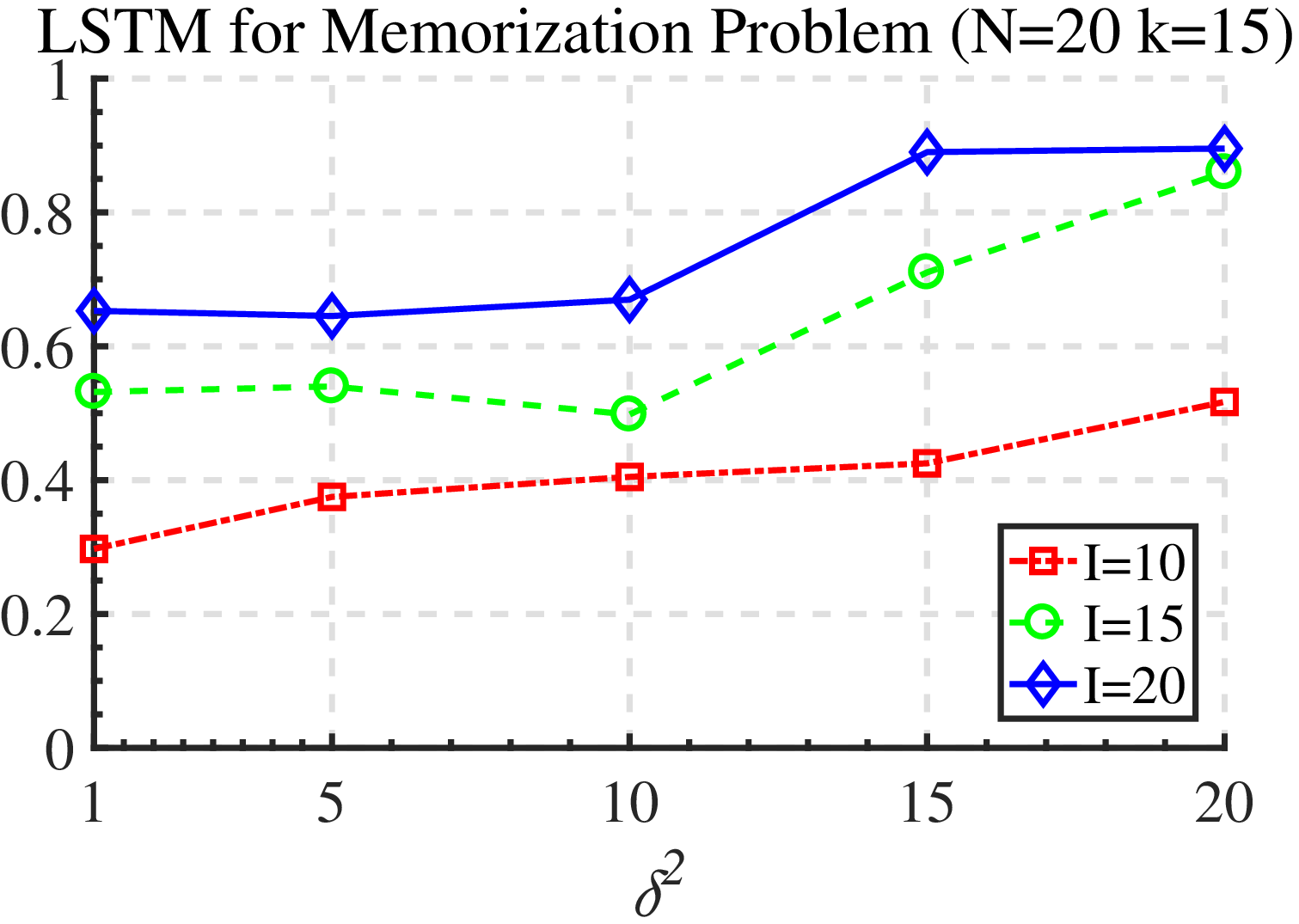} &
\includegraphics[width=.30\textwidth]{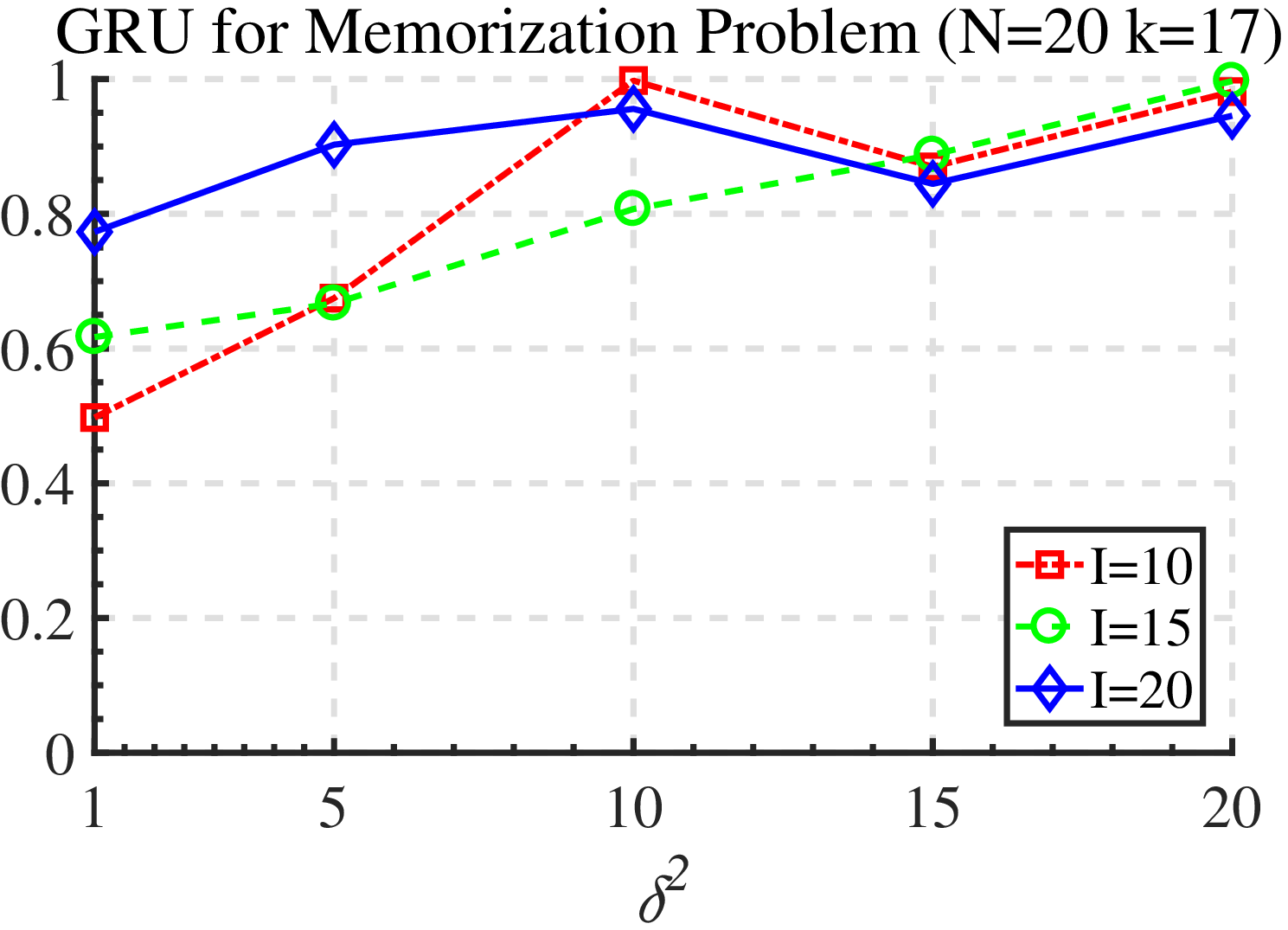}&
\includegraphics[width=.30\textwidth]{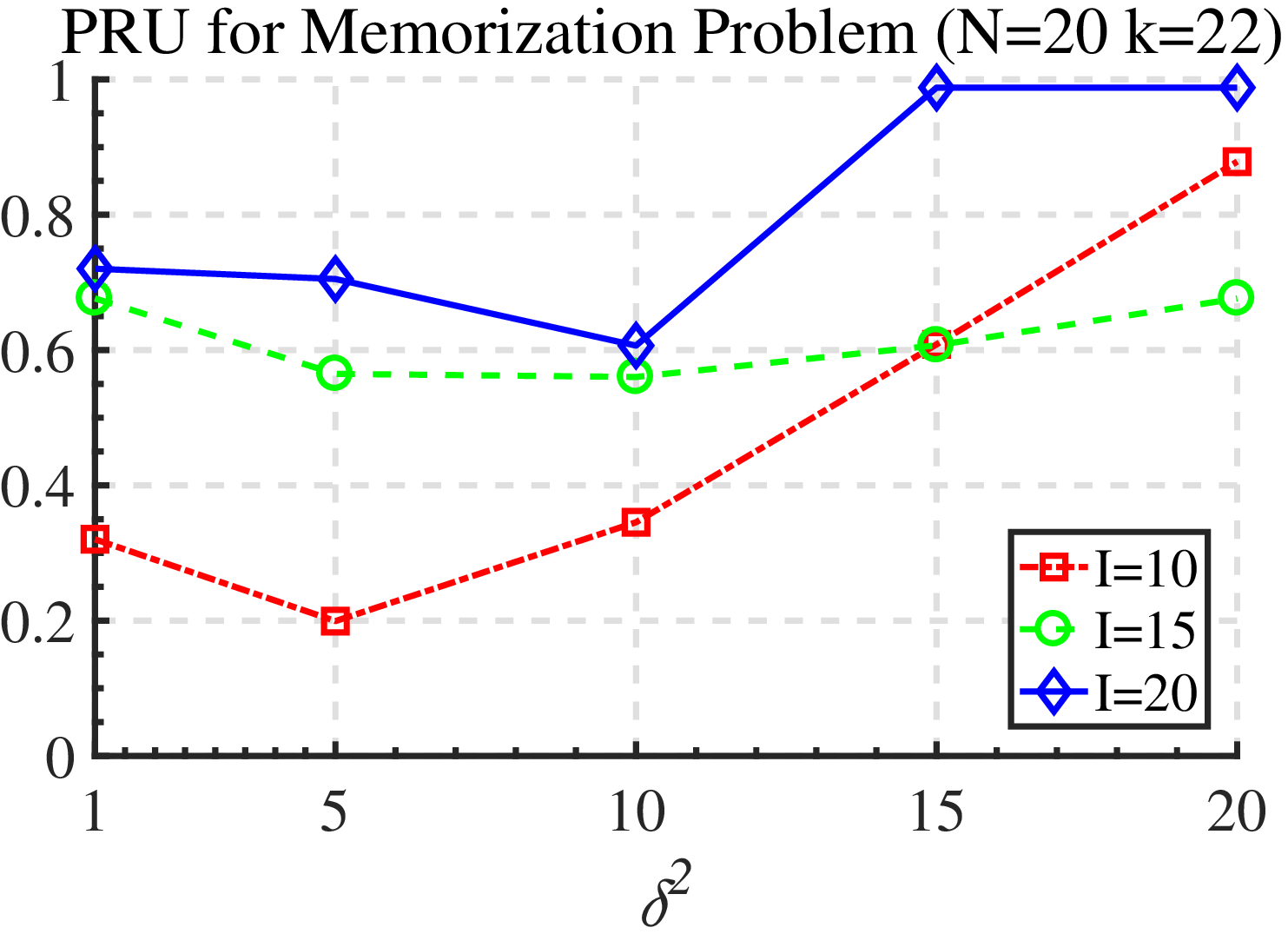}
\end{tabular}
\caption{\label{fig:problem_MP} MSE comparison of LSTM, GRU and PRU networks with varying parameters in Memorization Problem}
\end{figure*}

\begin{figure*}[htb!]\centering
\begin{tabular}{ccc}
\includegraphics[width=.30\textwidth]{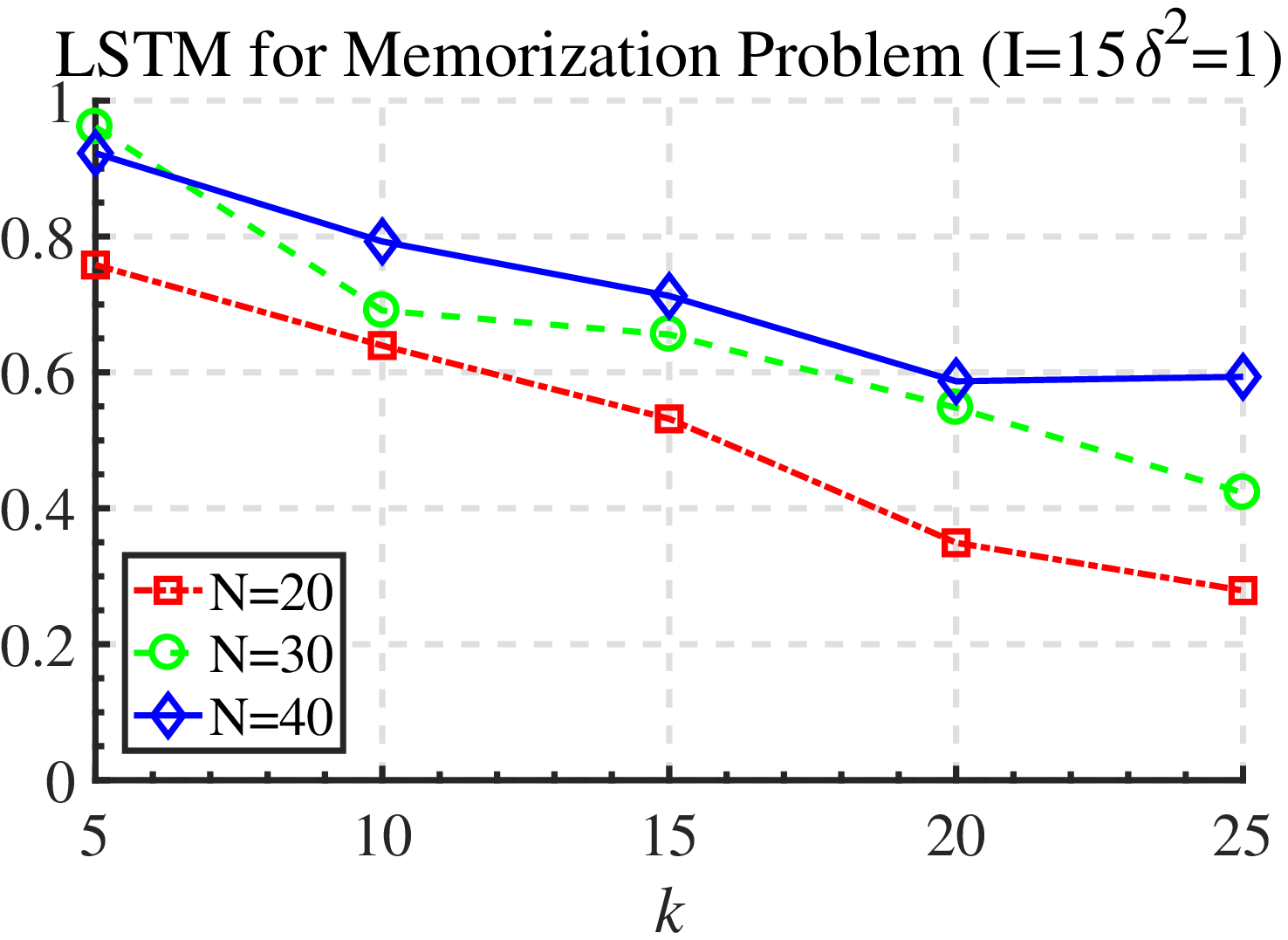} &
\includegraphics[width=.30\textwidth]{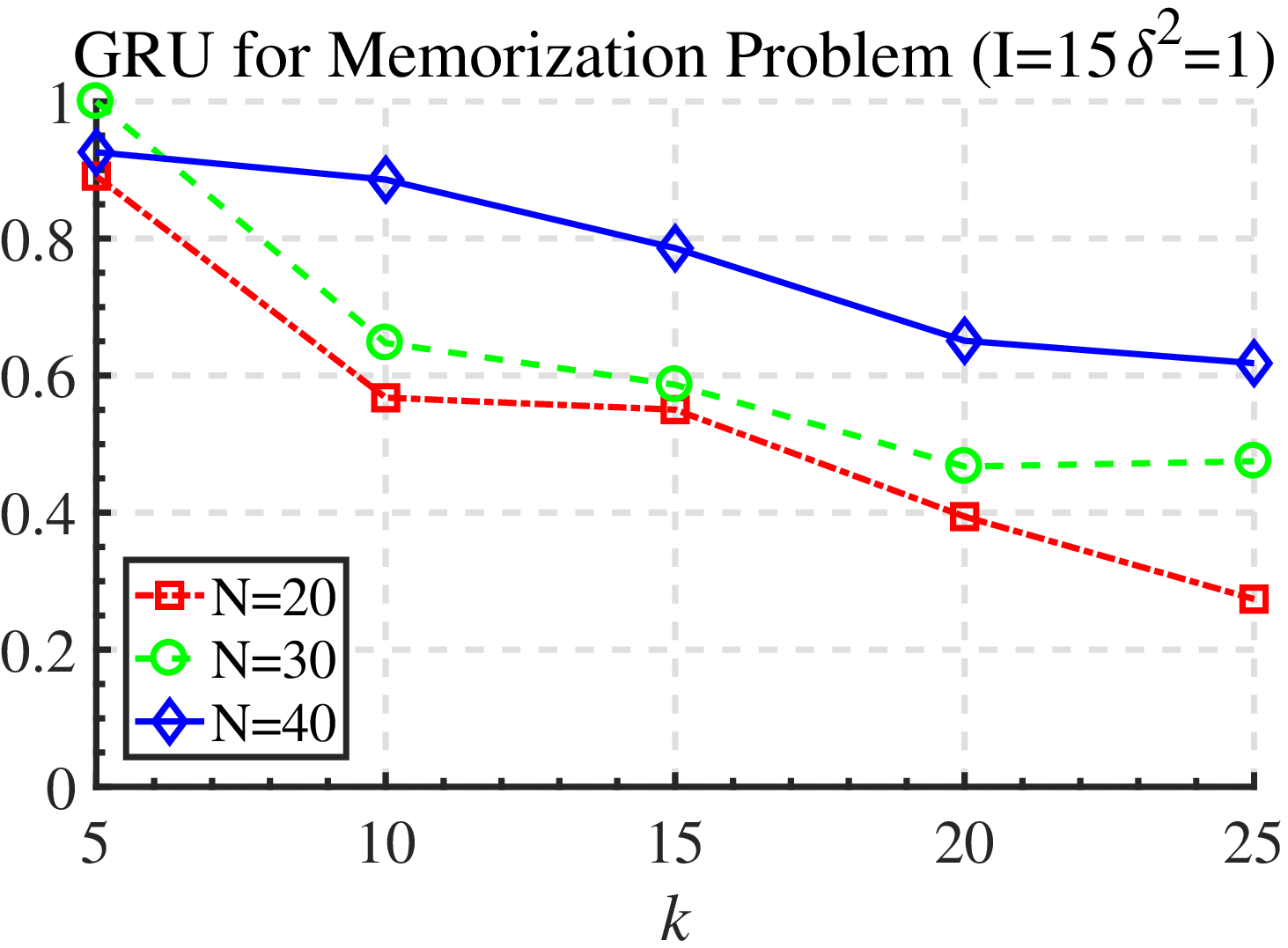} &
\includegraphics[width=.30\textwidth]{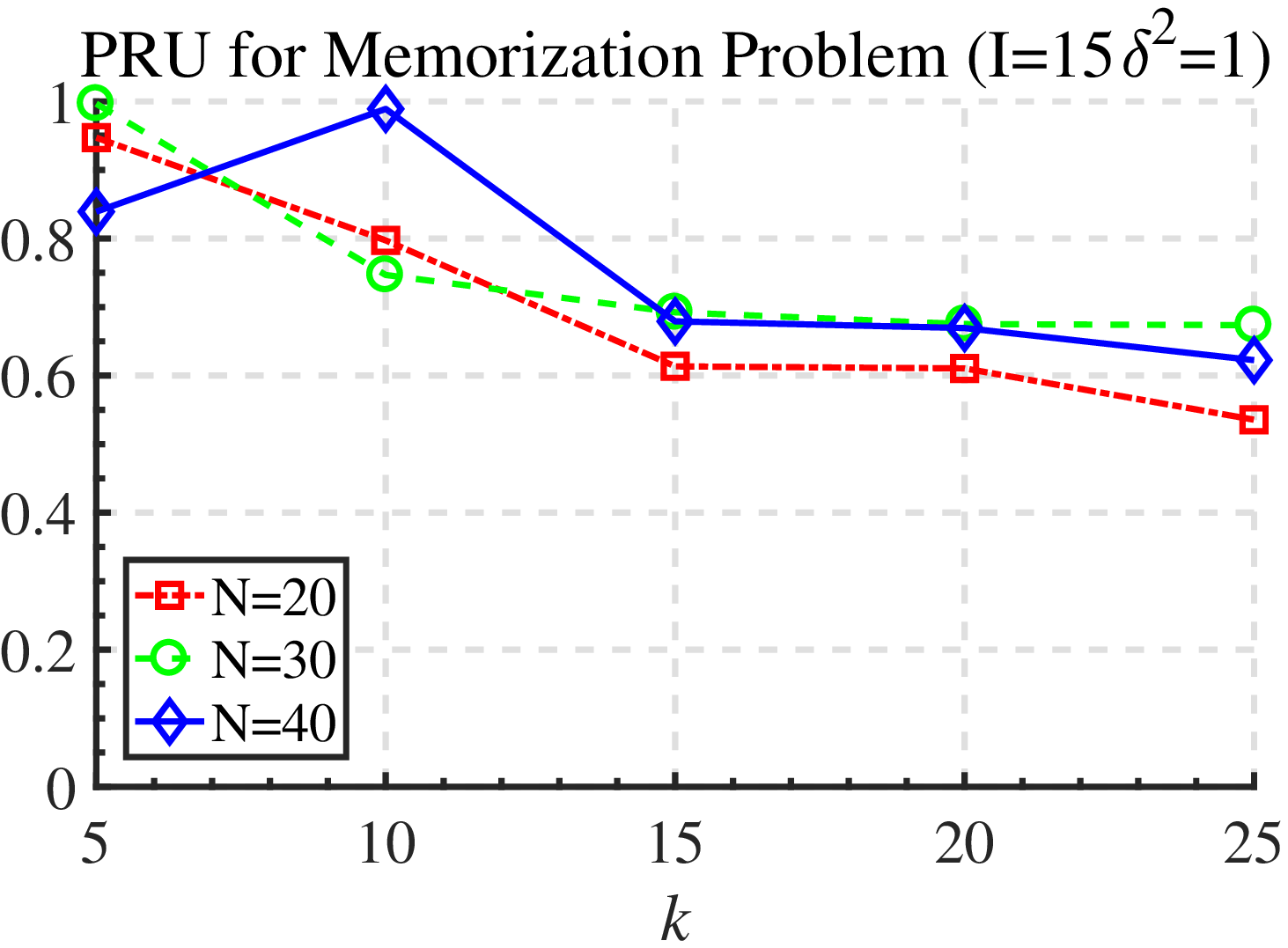}
\end{tabular}
\caption{\label{fig:state_MP} MSE comparison of LSTM, GRU and PRU networks with varying state-space  dimension $k$.}
\end{figure*}

\noindent \underline{\bf Results:} Results are obtained for LTSM, GRU and PRU under various problems settings $(I, N, \delta^2)$ and model state-space dimensions $k$.

Figure \ref{fig:overall_MP} shows the performance comparison of the three recurrent units with same state space dimension. It can be seen that the three units perform similarly, among which LSTM's performance is superior to the other two with same state space dimension. However, in this comparison, LSTM uses the most parameters. Figure \ref{fig:mpsame parameter} demonstrates the performances under the same number of parameters. It can be observed that PRU outperforms GRU, even catches up the performance of LSTM to a certain extent. In addition, with respect to any given parameter, the performance trends of the three units are identical.

Figure \ref{fig:problem_MP} shows how the performance of each unit is related to the problem parameters $I$, $N$, and $\delta^2$.  For every unit and a fixed state space dimension $k$, the following performance trend can be observed.
\begin{itemize}
\item The performance degrades with increasing $I$. That is, when the amount of targeted information increases, it becomes more difficult for the unit to memorize this information.
\item The performance degrades with increasing $N$. That is, over a long period of time, the units tend to forget the targeted information.
\item The performance degrades with increasing $\delta^2$. That is, when the interfering signal become stronger, it is more difficult to memorize the targeted information.
\end{itemize}

Figure \ref{fig:state_MP} shows how the performance of each unit varies with the state space dimension $k$. It is apparent from the figure that as the dimension of state space increases, the performance of each unit improves. This behaviour is sensible, since the role of the state variable in a recurrent unit may be intuitively understood as the ``container'' for ``storing'' information, and large state space would result in larger ``storage capacity''.

%From Table \ref{tab:time_MP},
From Table \ref{tab:time_MP} (measured at $k=15$ and $I=15$),  it can be observed that PRU has lowest time complexity, significantly below GRU and LSTM. This is a direct consequence of PRU's structural simplicity.

\begin{table}[ht!]
\caption{\label{tab:time_MP}Time cost per epoch in Memorization Problem, $k=3$, $I=2$}
\centerline{
\begin{tabular}{|@{\hspace{0.2cm}}c|@{\hspace{0.2cm}}c|@{\hspace{0.2cm}}c|@{\hspace{0.2cm}}c|@{\hspace{0.2cm}}c|@{\hspace{0.2cm}}c|}
\hline
 $N$ & 20 & 30 & 40 & 50 & 60 \\
\hline
LSTM & 40.445 & 60.575 & 80.653 & 102.435 & 125.245 \\
\hline
GRU & 30.355 & 45.863 & 60.524 & 76.328 & 93.558 \\
\hline
PRU & 20.472 & 31.791 & 43.434 & 55.131 & 70.244 \\
\hline
\end{tabular}
}

\end{table}

\subsection{Adding Problem}

To describe the Adding Problem, let ${\cal M}_{\rm add}$ be an ``adding machine'', which is a function mapping a length-$N$ sequence $(x_1, x_2, \ldots, x_N)$ to a real number. In particular, each $x_t$, $t=1, 2, \ldots, N$, is a vector in ${\mathbb R}^2$, and we may write $x_t$ as $(x_t[1], x_t[2])^T$. At each $t$,  $x_t[1]$ is a random value drawn independently from the zero-mean Gaussian distribution with variance $\delta^2$; and in the sequence $(x_1[2], x_2[2], \ldots, x_N[2])$, there are exactly two $1$'s, the locations of which are randomly assigned; the remaining values of the sequence all are equal to $0$.
The behaviour of the adding machine is given by
\[
{\cal M}_{\rm add}(x_1, x_2, \ldots, x_N):= \sum\limits_{t=1}^N x_t[1]\cdot x_t[2].
\]
The objective of the Adding Problem is then to train a model that simulates the behaviour of ${\cal M}_{\rm add}$. Obviously, the Adding Problem is parametrized by the pair $(N, \delta^2)$.  Intuitively,   the Adding Problem demands higher ``memorization capacity'' than the Memorization Problem, since only counting the locations of the two $1$'s in the second component the input sequence, there are ${N \choose 2}$ possibilities.

\noindent \underline{\bf Modelling:} Under a recurrent network model, it is natural to take input space ${\pazocal X}={\mathbb R}^2$
and  output space ${\pazocal Y}={\mathbb R}$. Except at the final time $t=N$, the output $y_t$ is discarded, and final output $y_{N}$ is used to simulate the output ${\cal M}_{\rm add}(x_1, x_2, \ldots, x_{N})$ of the adding machine.

\noindent \underline{\bf Datasets:} For each problem setting $(N, \delta^2)$, we generate 2000 training examples and 400 testing examples according to the specification of the problem.

\noindent \underline{\bf Training:}  The training of each model is performed by optimizing the MSE between the $y_N$ and
${\cal M}_{\rm add}(x_1, x_2, \ldots, x_{N})$. A mini-batched SGD method is used for optimization, where we use the same set of training parameters as those in the Memorization Problem, except that the batch size is chosen as $50$.

\noindent \underline{\bf Evaluation Metrics:} MSE is used as the evaluation metric, and the same averaging process as that for the Memorization Problem is applied.

%The parameter settings of experiments for Adding Problem are given below:
%\begin{itemize}
%\item $N = 1, 2, 3, 4, 5, 6, 7, 8, 9, 10$
%\item $k = 1, 2, 3$
%\item $\sigma^2=1, 4, 9$.
%
%\end{itemize}
%The graph below depicts the experimental test error changes over different parameters.
%
% \input{figs_adding}

%\begin{table}[h]
%\centering
%\begin{tabular}{|c|c|c|c|c|c|}
%\hline
% $N$ & 2 & 4 & 6 & 8 & 10 \\
%\hline
%LSTM & 0.4678 & 0.8097 & 1.1826 & 1.5450 & 2.0387 \\
%\hline
%GRU & 0.4346 & 0.7270 & 0.9954 & 1.3106 & 1.5756 \\
%\hline
%PRU & 0.3191 & 0.5822 & 0.7367 & 0.9380 & 1.2037 \\
%\hline
%\end{tabular}
%\caption{\label{tab:time_AP}Time cost per epoch for Adding Problem, k=3}
%\end{table}
\begin{figure*}[htb!]\centering
\begin{tabular}{ccc}
\includegraphics[width=.30\textwidth]{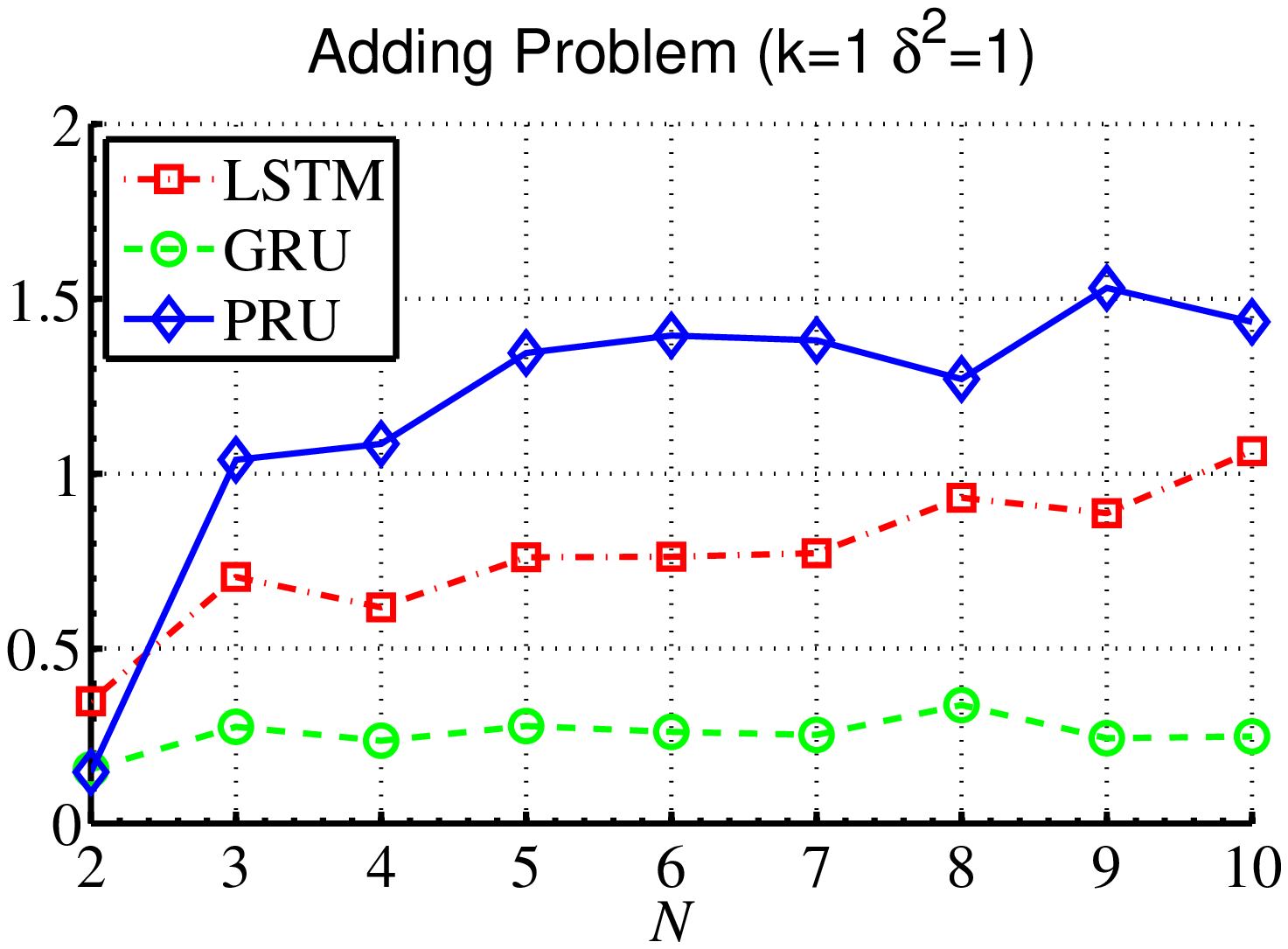} &
\includegraphics[width=.30\textwidth]{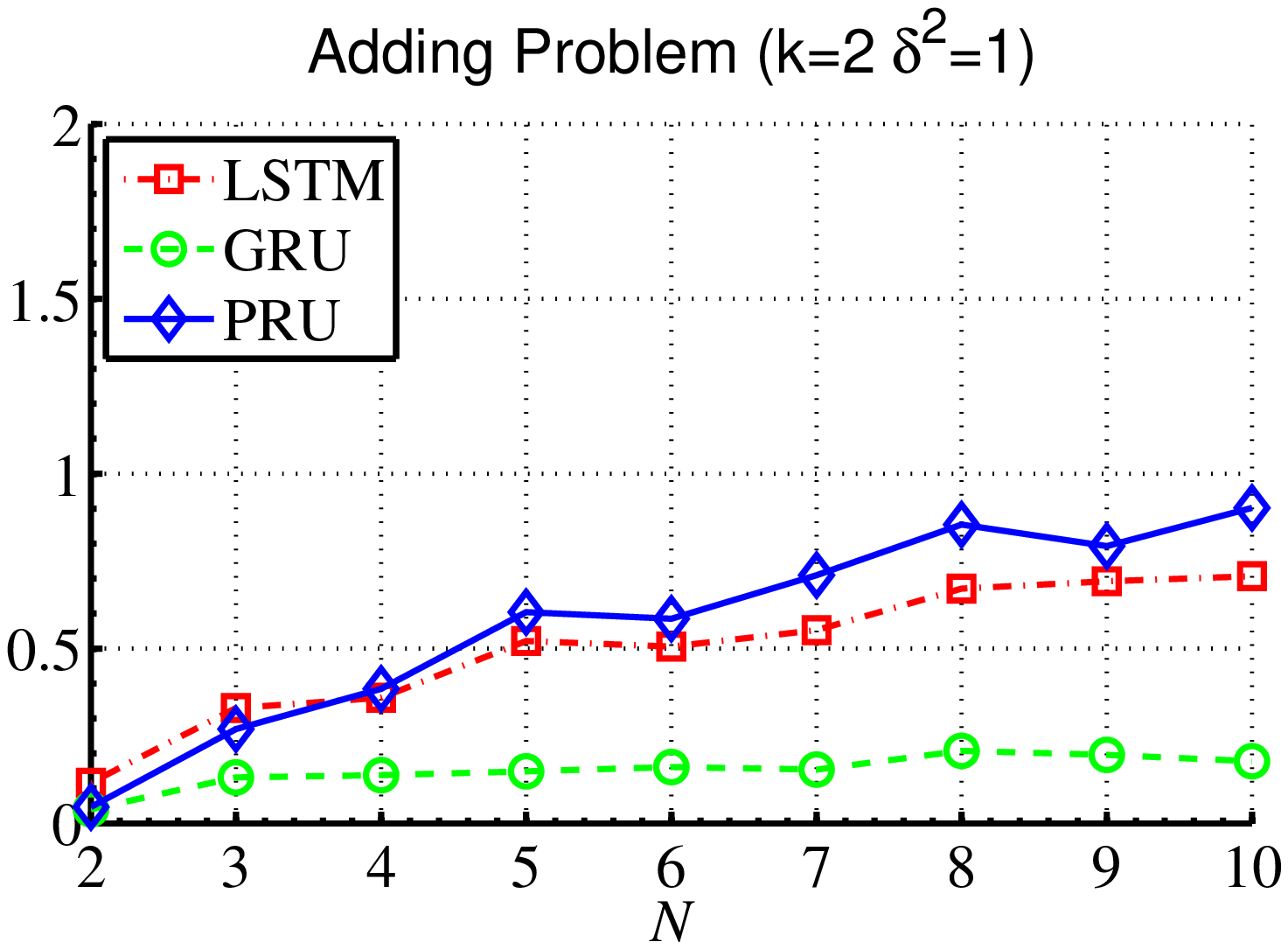} &
\includegraphics[width=.30\textwidth]{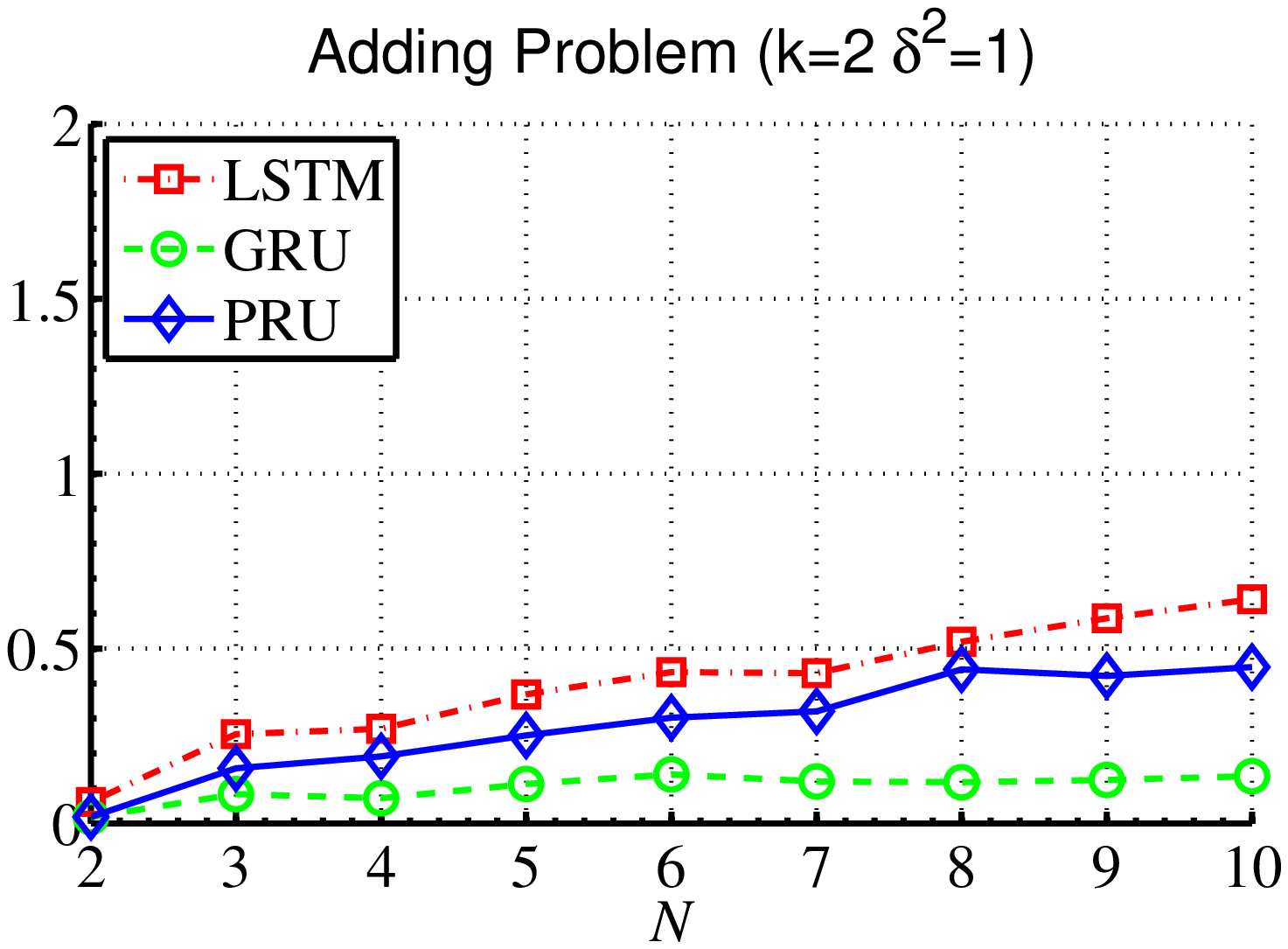}
\end{tabular}
\caption{\label{fig:overall_AP} MSE comparison of LSTM, GRU and PRU in Adding Problem.}
\end{figure*}

\begin{figure*}[htb!]\centering
\begin{tabular}{ccc}
\includegraphics[width=.30\textwidth]{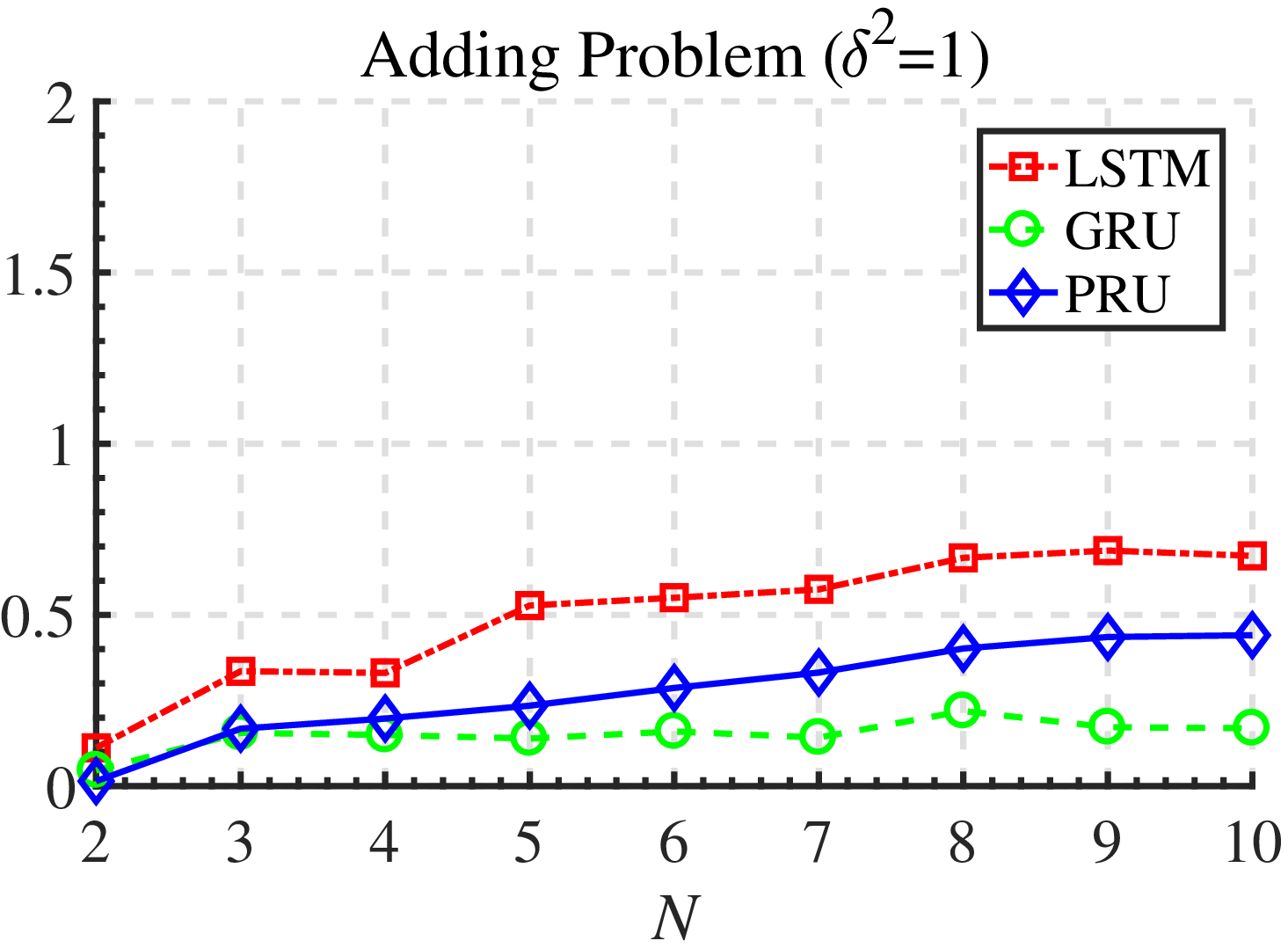} &
\includegraphics[width=.30\textwidth]{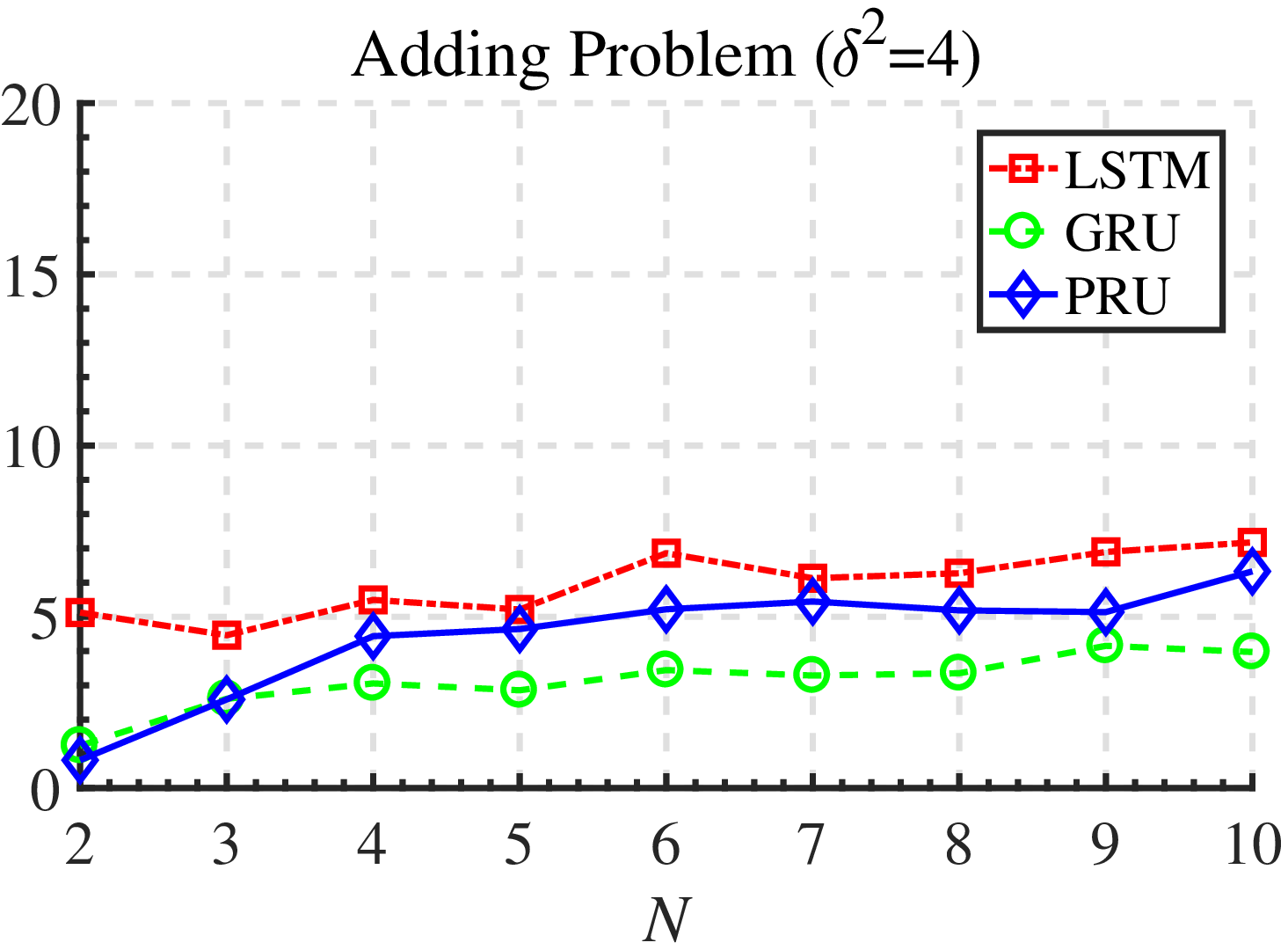} &
\includegraphics[width=.30\textwidth]{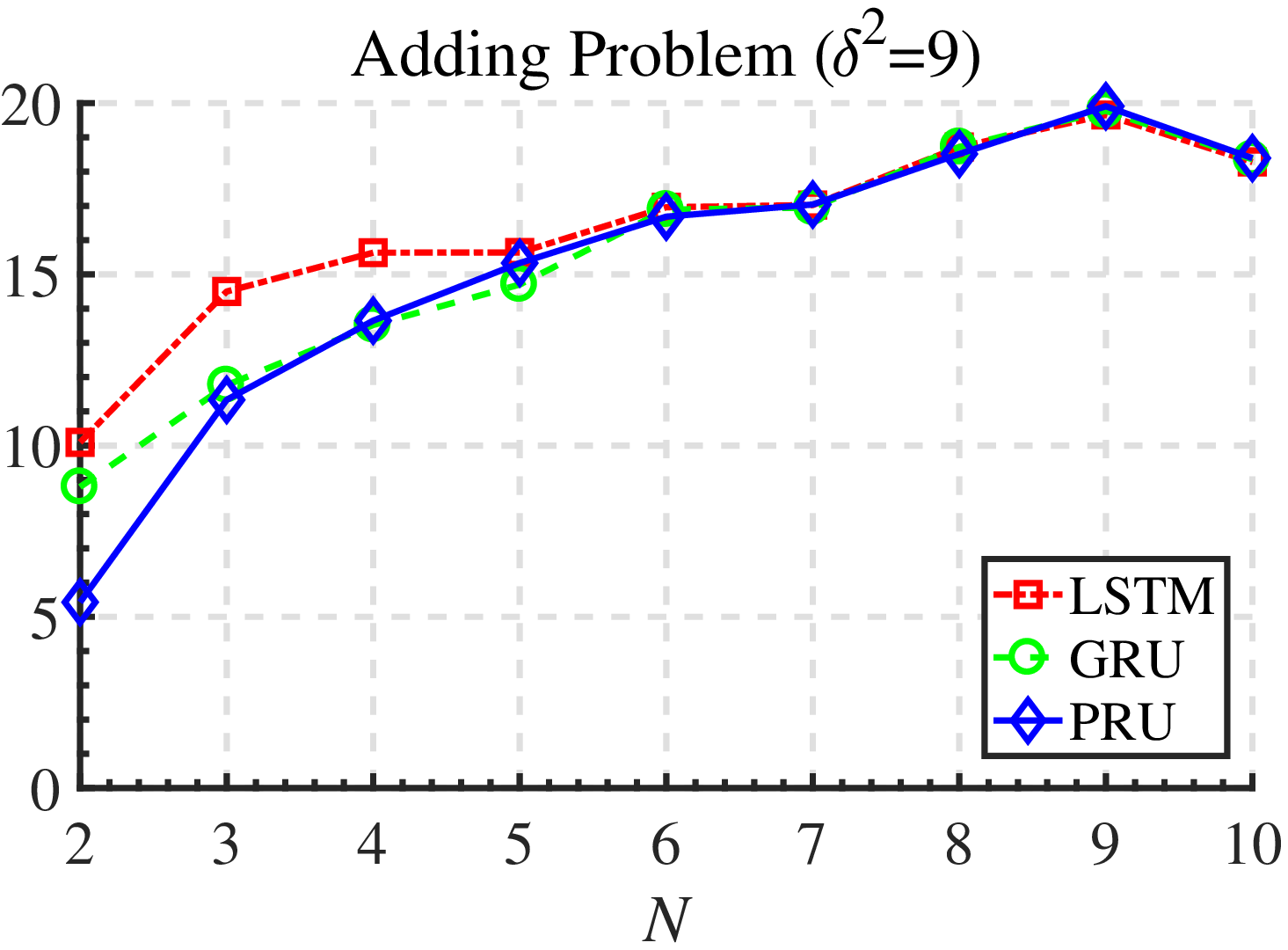}
\end{tabular}
\caption{\label{fig:same parameter} MSE comparison of LSTM, GRU and PRU in Adding Problem with same number of parameters.}
\end{figure*}

\begin{figure*}[htb!]\centering
\begin{tabular}{ccc}
\includegraphics[width=.30\textwidth]{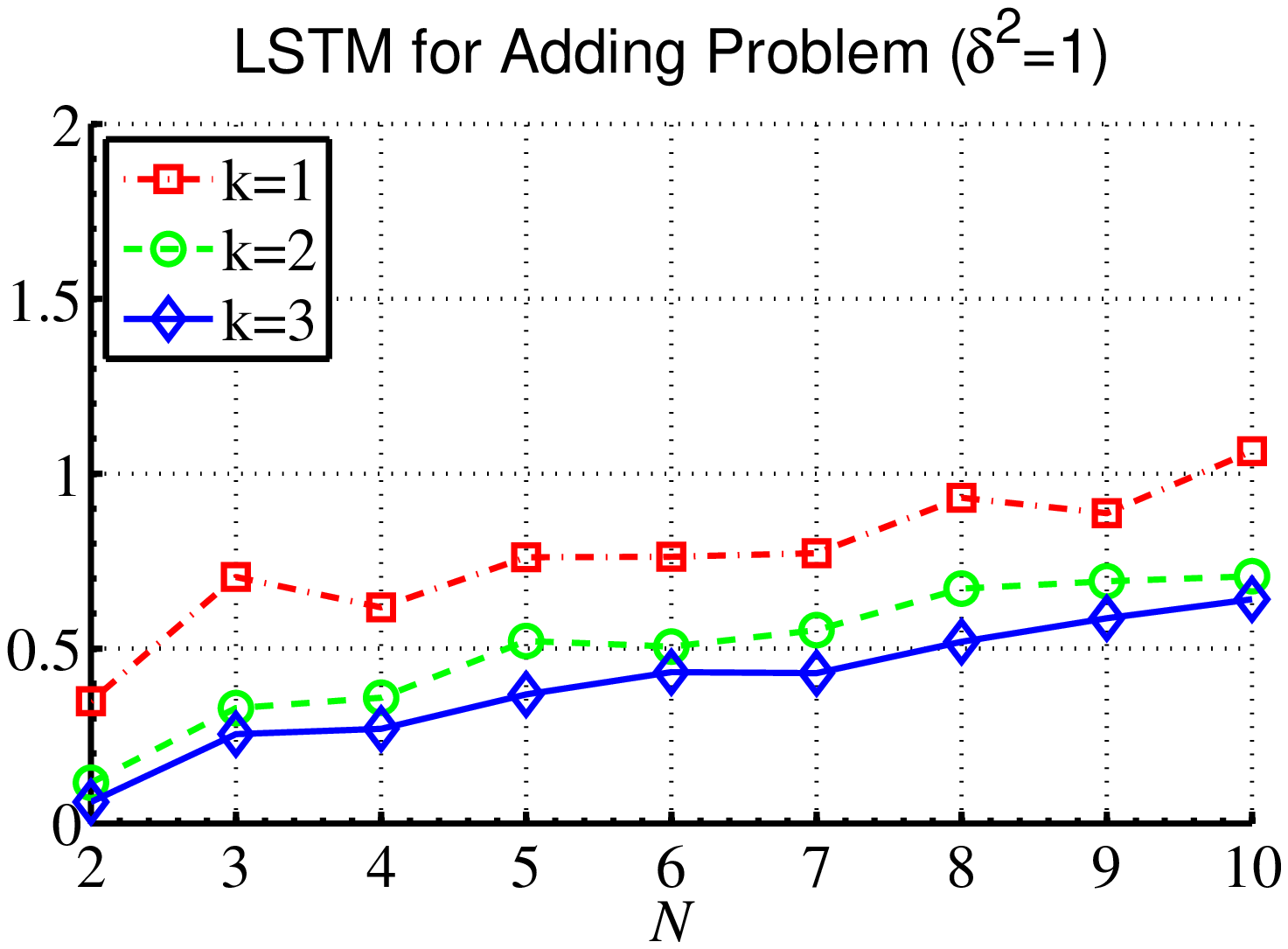} &
\includegraphics[width=.30\textwidth]{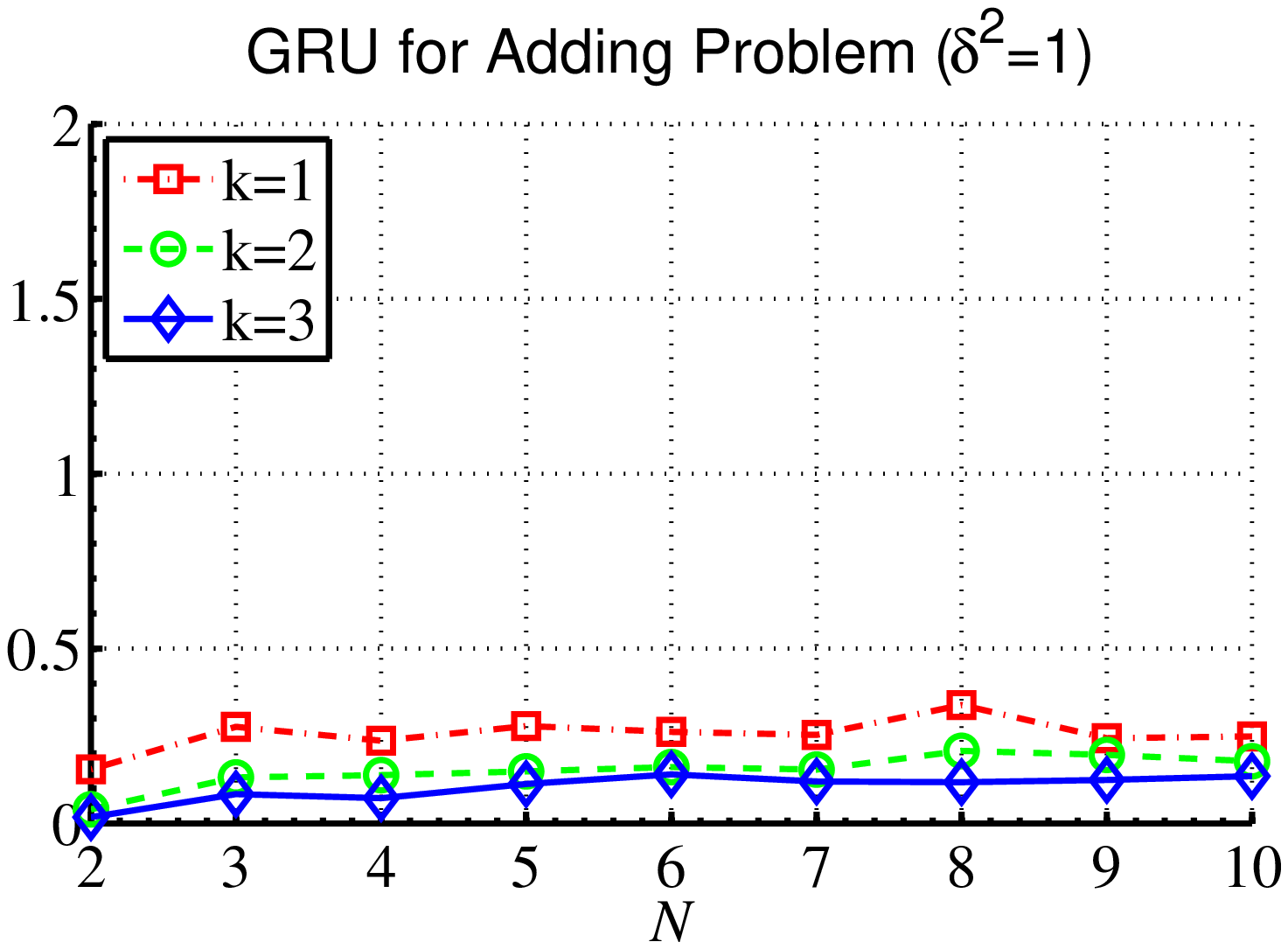} &
\includegraphics[width=.30\textwidth]{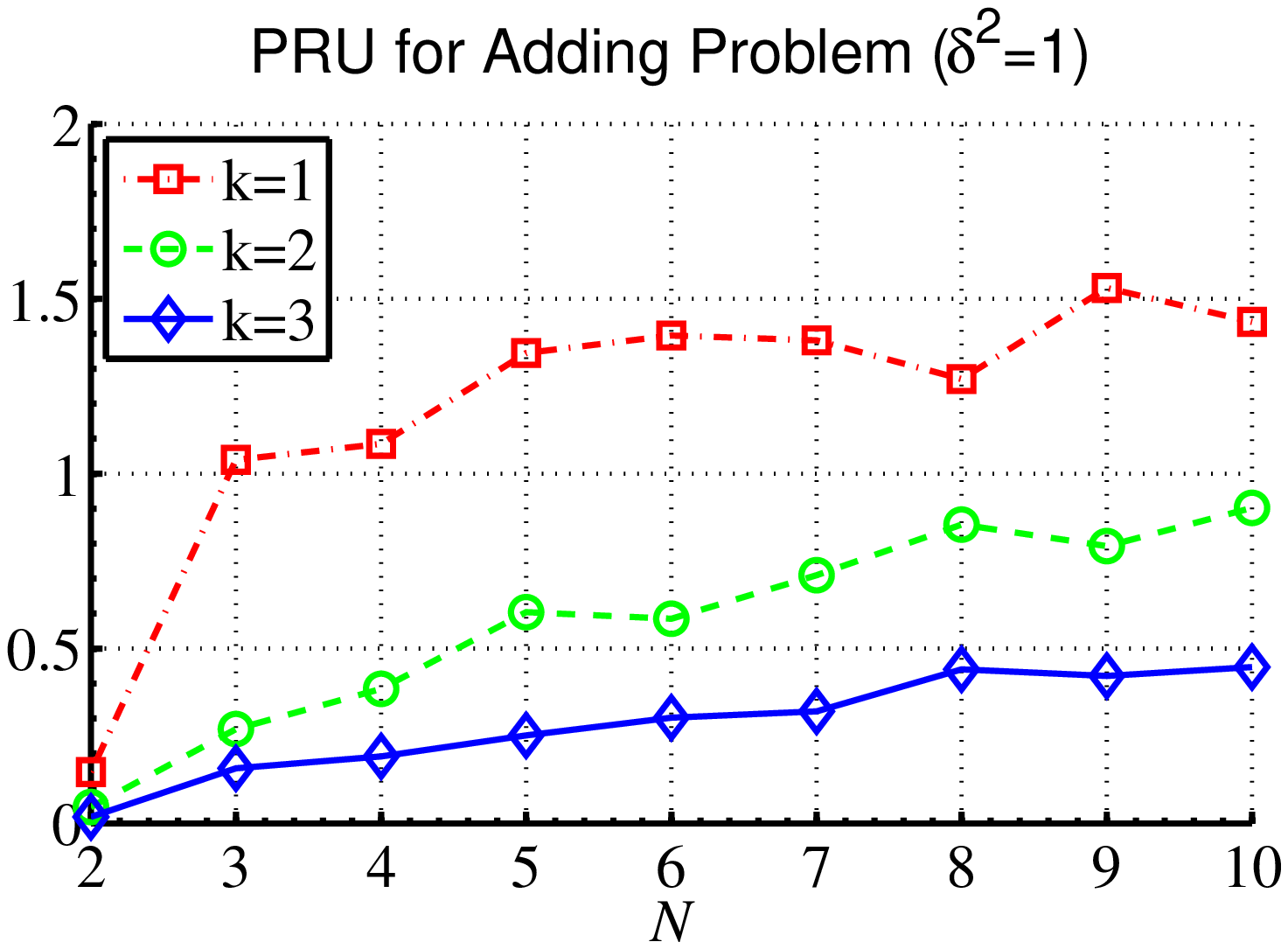} \\
\includegraphics[width=.30\textwidth]{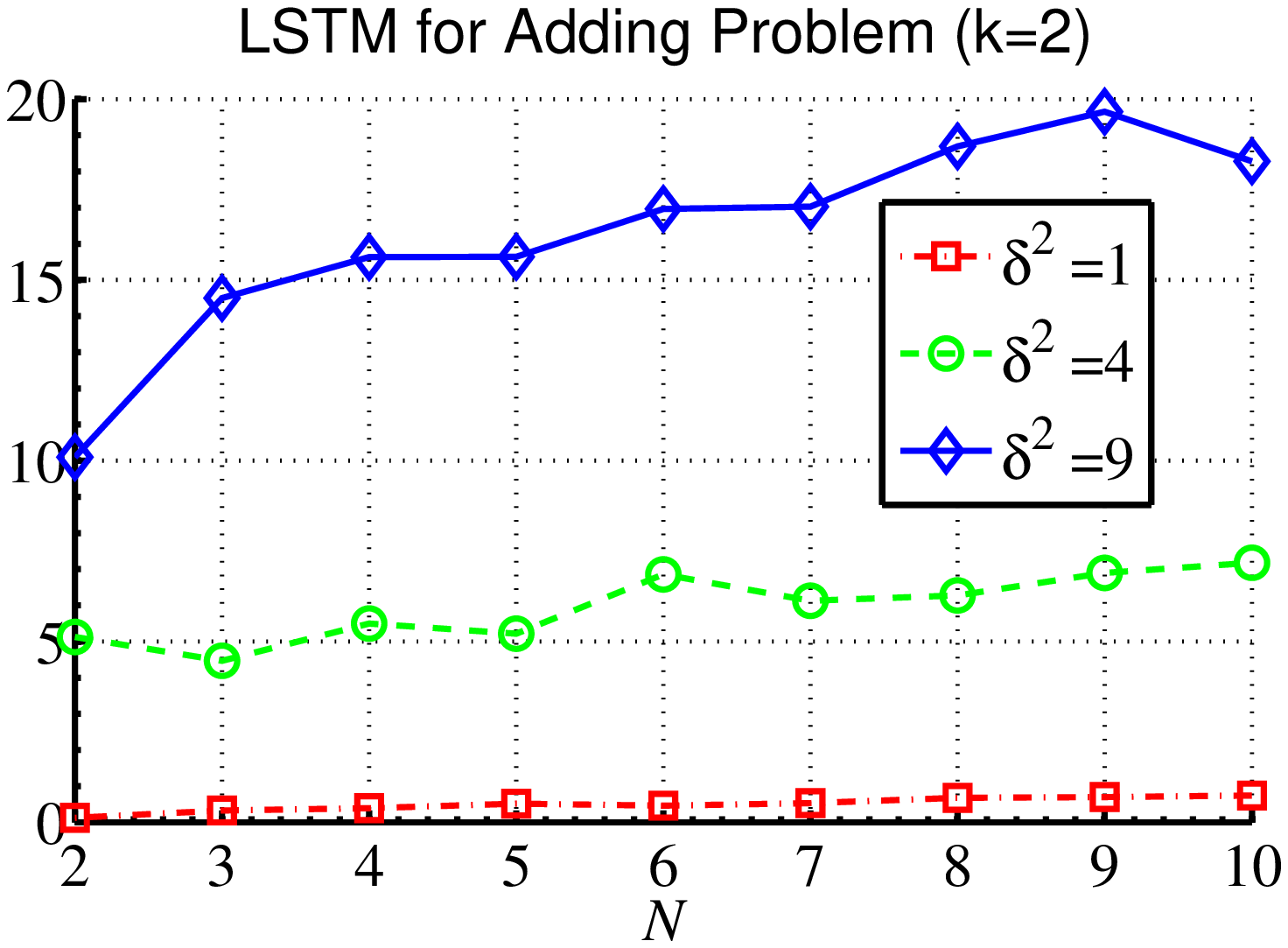} &
\includegraphics[width=.30\textwidth]{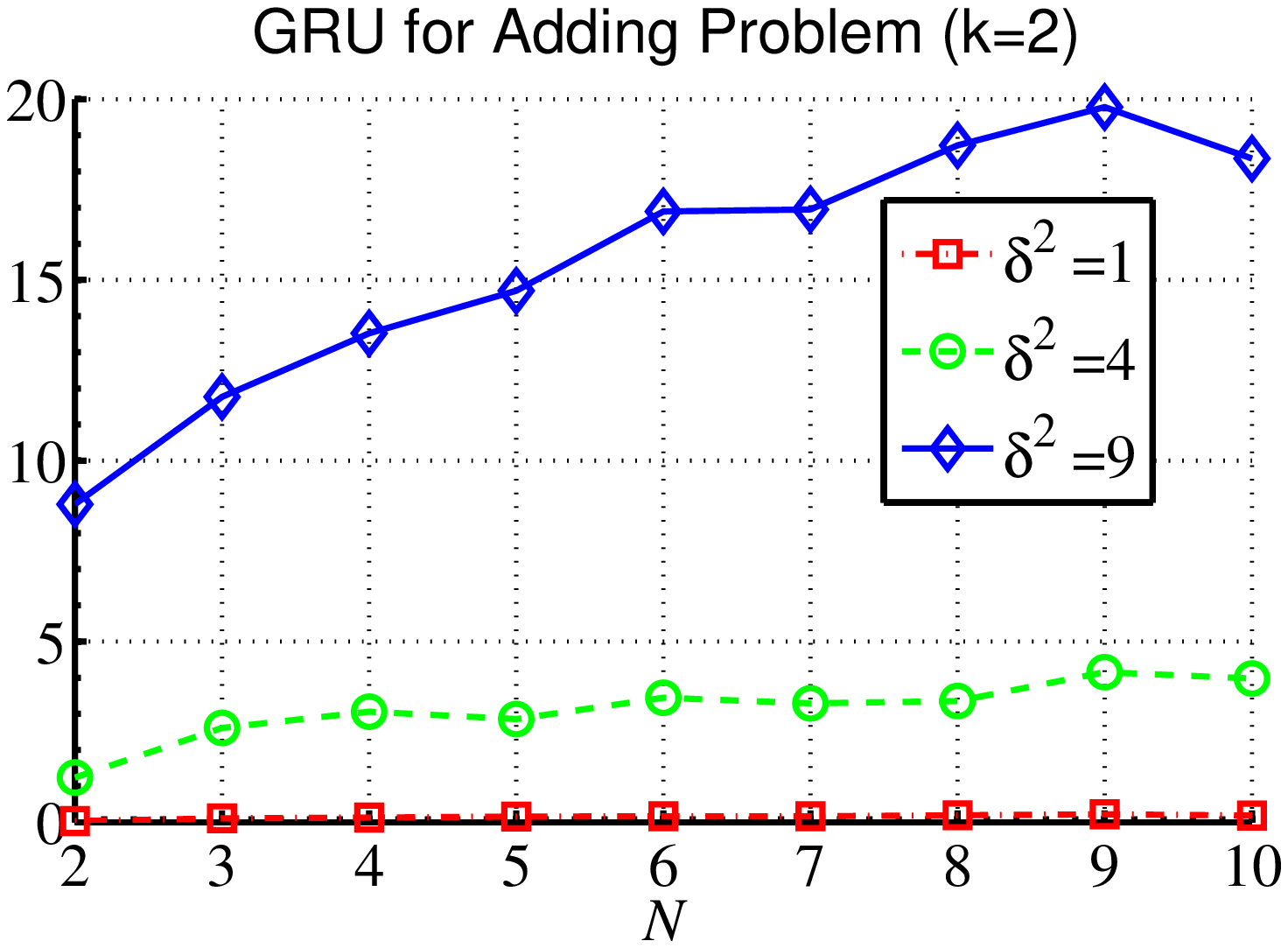} &
\includegraphics[width=.30\textwidth]{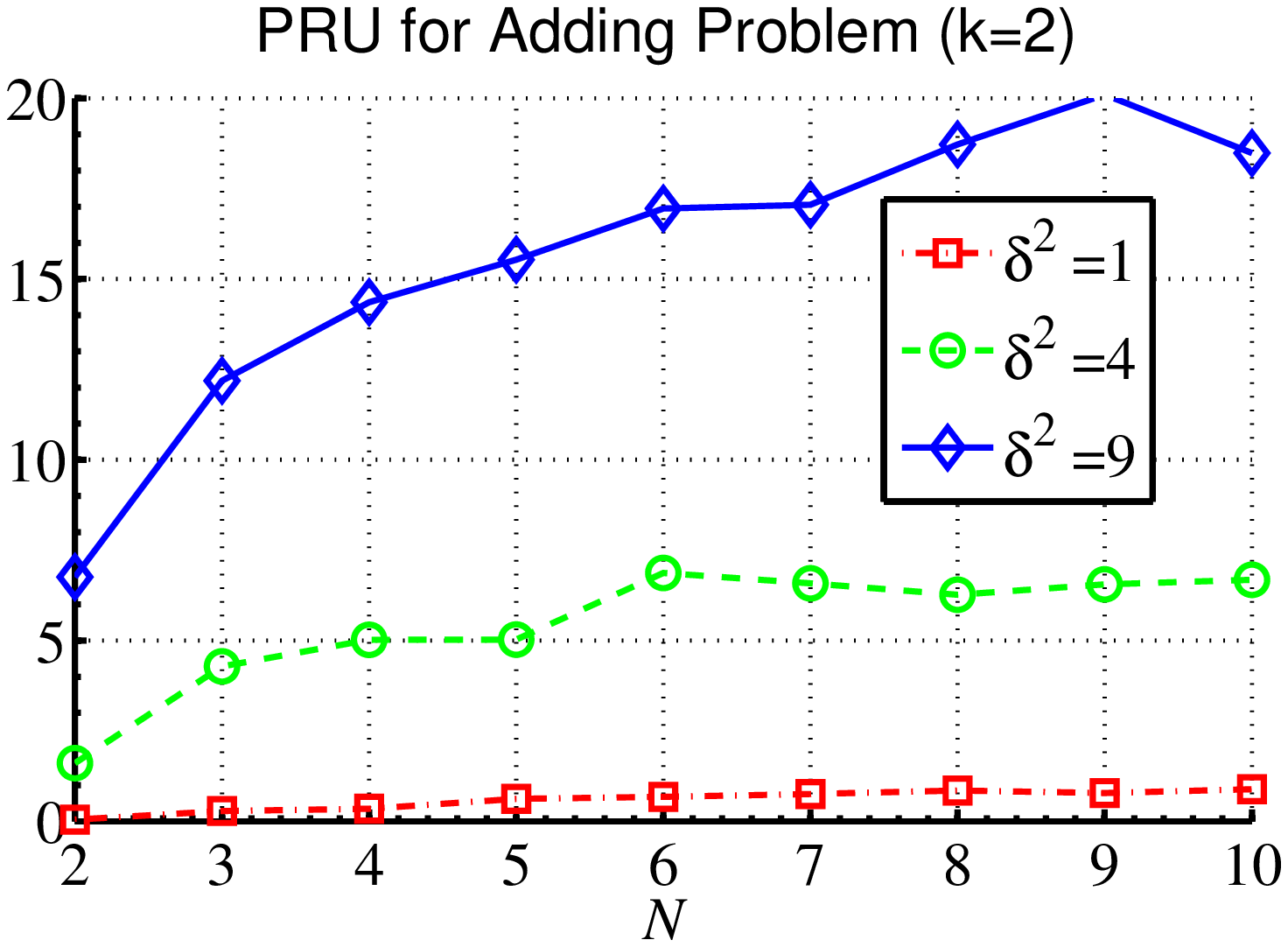}
\end{tabular}
\caption{\label{fig:trends_AP} MSE comparison of LSTM, GRU, and PRU in Adding Problem under varying experiment settings. The MSE value has been normalized over $\delta^2$.}
\end{figure*}

\noindent \underline{\bf Results:} Figure \ref{fig:overall_AP} shows the performance comparison of LSTM, GRU and PRU in the Adding Problem with the same state space dimension, Figure \ref{fig:same parameter} shows the performance comparison of the three recurrent units with the same number of parameters and Figure \ref{fig:trends_AP} shows the performance trend of each of the three units with respect varying parameters. In Adding Problem, overall the three units perform comparably, GRU superior to the other two units. It is worth noting in Figure \ref{fig:overall_AP}, with low state-space dimension ($k=1$), PRU appears under-perform LSTM. But as the state-space dimension increases, PRU catches up (at $k=2$) and even out-performs LSTM (at $k=3$). This may be explained as follows. First the Adding Problem demands higher ``memorization capacity''. But as we discussed earlier, PRU uses the Type-II representations, which may need larger state-space for the same representation power. In Figure \ref{fig:same parameter}, it can be seen that the performance difference between PRU and GRU is narrowing with same number of parameters, additionally, under certain parameter settings, the performance between PRU and GRU is almost with no difference even the same.

These results also suggest that the three studied units all have identical performance trends with respect to state-space dimension or any given problem parameter. Conclusions similar to those in the Memorization Problem may be obtained. The time complexity of PRU is also the lowest among the three for the Adding Problem(table below, measured at $k=3$).
%(Table \ref{tab:time_AP}).

\begin{table}[h]
\caption{\label{tab:time_AP}Time cost per epoch for Adding Problem, $k=3$}
\centerline{
\begin{tabular}{|@{\hspace{0.2cm}}c|@{\hspace{0.2cm}}c|@{\hspace{0.2cm}}c|@{\hspace{0.2cm}}c|@{\hspace{0.2cm}}c|@{\hspace{0.2cm}}c|}
\hline
 $N$ & 2 & 4 & 6 & 8 & 10 \\
\hline
LSTM & 0.4678 & 0.8097 & 1.1826 & 1.5450 & 2.0387 \\
\hline
GRU & 0.4346 & 0.7270 & 0.9954 & 1.3106 & 1.5756 \\
\hline
PRU & 0.3191 & 0.5822 & 0.7367 & 0.9380 & 1.2037 \\
\hline
\end{tabular}
}

\end{table}

%
%\begin{itemize}
%\item From the experiment results of Adding Problem, we can get some similar conclusion with Memorization problem, for Adding problem, the performance of the three recurrent networks is also comparable. In general, GRU has the best performance.
%\item For a fixed hidden state dimension, the test error increases with the increasing of the sequence length, when the sequence length is $N$, Adding Problem faces with ${N \choose 2}$ modes to adding the two components, so the increase of $N$ means that the recurrent unit need to ``Memory'' more modes for addition.
%\item For a fixed length of the sequence,the performance of recurrent network is getting better with the increase of $k$. Similar to the Memorization problem, as the noise variance$\delta^2$ increasing, the normalize test error increases.
%\end{itemize}

\vspace{-0.2cm}
\subsection{Character Prediction Problem}

%For the Memorization problem and the Adding problem, all experimental data is generated by our own. We investigated the empirical performance of recurrent units for The Sequential Character Prediction Problem. We conducted a set of proof-of-concept experiments, the goal is not to compete with previous work or to find the best performing model under a specific hyperparameter setting. Rather, we investigate how the three recurrent units perform under different settings.

Let ${\cal M}_{\rm char}$ be a ``character-prediction machine'', which takes an input sequence $(x_1, x_2, \ldots, x_N)$
of arbitrary length $N$ and produces an output sequence of the same length. The input sequence is fed to the machine one symbol per time unit, and at each time $t$, the machine is characterized by a function ${\cal M}_{\rm char}^t$ defined by
\[
{\cal M}_{\rm char}^t(x_1, \ldots, x_t):= x_{t+1}.
\]
That is, for every input sequence, the output of the machine is the input sequence shifted in time. Here each symbol $x^t$ is a character in a $K$-character alphabet. Each character in the alphabet is represented by a length-$K$ {\em one-hot} vector.
%Thus, the input space of
%
% In particular, each $x_t$, $t=1, 2, \ldots, N$, is a $K$-dimensional ``{\textit one-hot}'' representation in ${\mathbb R}^K$.
%The behaviour of the character-prediction machine is given by
%\[
%{\cal M}_{\rm char}(x_1, \ldots, x_i):= (x_{i+1}).
%\]
The objective of the Character Prediction Problem is then to train a model that simulates the behaviour of ${\cal M}_{\rm char}$.

\noindent \underline{\bf Modelling:} Naturally, both  ${\pazocal X}$ and ${\pazocal Y}$ are taken as ${\mathbb R}^K$
 in the models. The output $y_t$ is computed by a soft-max classifier.

\noindent \underline{\bf Dataset:} A Shakespeare drama dataset\footnote{https://github.com/karpathy/char-rnn}
is used in this experiment, where each sentence is taken as an input sequence.  The dataset consists of 1,115,393 occurrences of characters from an alphabet of size $64$, where 90\% of the sentences are used for training set and the rest is held out for testing.

\noindent \underline{\bf Training and Evaluation:}  The objective of this problem is to minimize the (expected) cross entropy loss (CEL)
\[
{\cal E}_{\rm CEL}(\theta)= -{\mathbb E}\left(
\frac{1}{N}
\sum_{t=1}^N\sum_{i=1}^{K}{\cal M}_{\rm char}^t(x_1, \ldots, x_t)[i]\log y_t[i]
\right)
\]
where we have used $v[i]$ to denote the $i^{\rm th}$ component of vector $v$, and used $N$ to denote the length of the input sequence.
 %$y_t$ is the output at the time step $t$ and $p_t$ is the target output.
 Mini-batched SGD with Adadelta dynamic learning rate \cite{zeiler2012adadelta} is used for optimization. All parameters are randomly initialized in $[-0.1, 0.1]$, and the base learning rate is set to 0.8. CEL is used to evaluate the models.

%
% We use one layer recurrent network, all the parameters is initialized uniformly in range [-0.1, 0.1], and we use mini-batch stochastic gradient descent with batch size is 200 and Adadelta \cite{zeiler2012adadelta}optimization algorithm with base learning rate 0.8.

%\noindent \underline{\bf Evaluation Metrics:} The cross entropy loss is used the evaluation metric, the experiment results gets from the hidden state dimension takes value in $\{64, 96, 128\}$,

%The table below shows the test cross-entropy loss under different parameter settings.
%\begin{table}[!hbp]
%\centering
%\begin{tabular}{|c|c|c|c|}
%\hline
% Hidden state dimension & LSTM & GRU & PRU \\
%\hline
%64 & 1.289 & 1.226 & 1.227 \\
%\hline
%96 & 1.225 & 1.216 & 1.188 \\
%\hline
%128 & 1.194 & 1.192 & 1.162 \\
%\hline
%\end{tabular}
%\caption{\label{tab:cel}Cross-entropy loss for Character Prediction}
%\end{table}

\begin{figure}[t]\centering
\begin{tabular}{ccc}
\includegraphics[width=.3\textwidth]{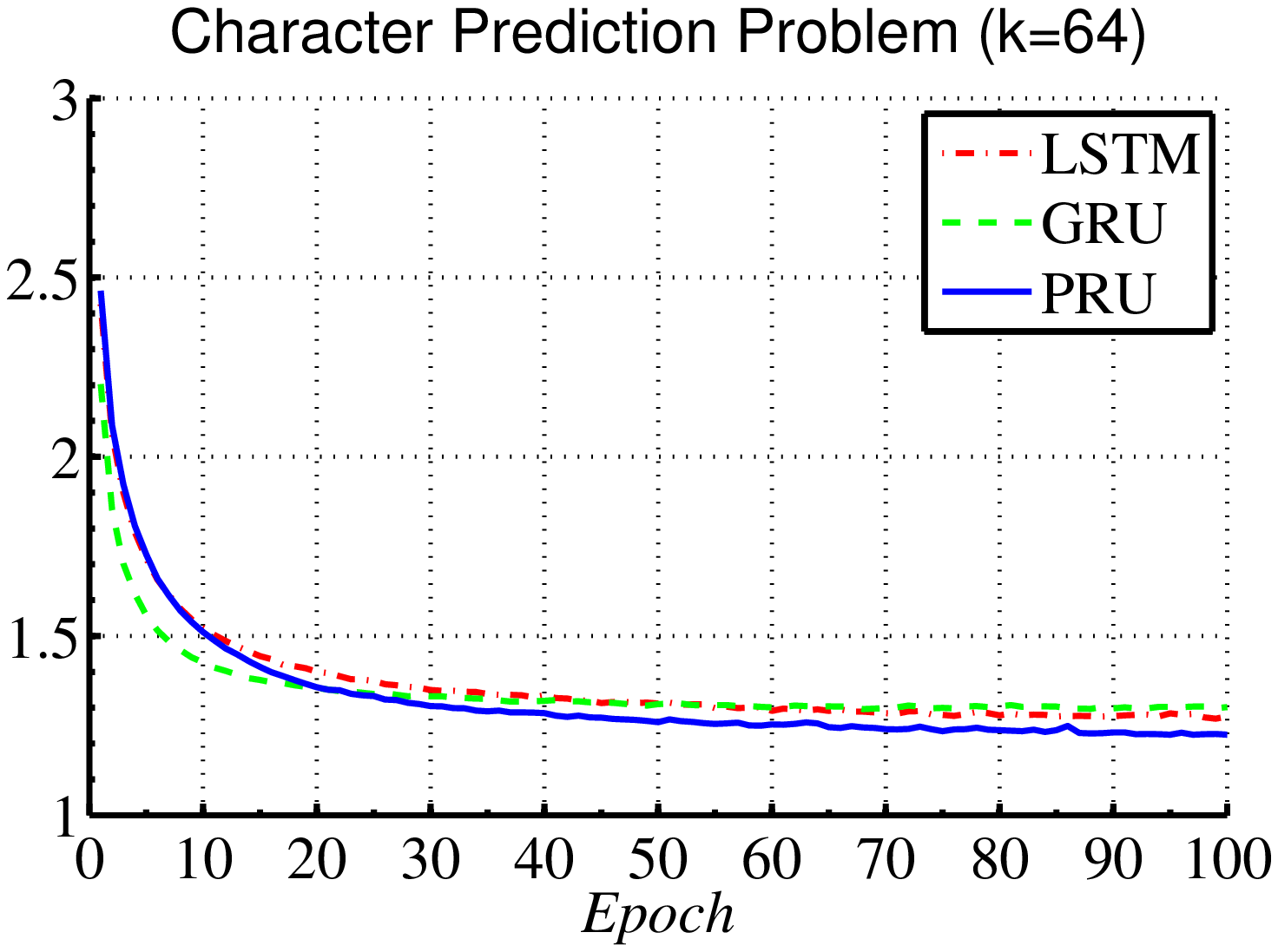} &
\includegraphics[width=.3\textwidth]{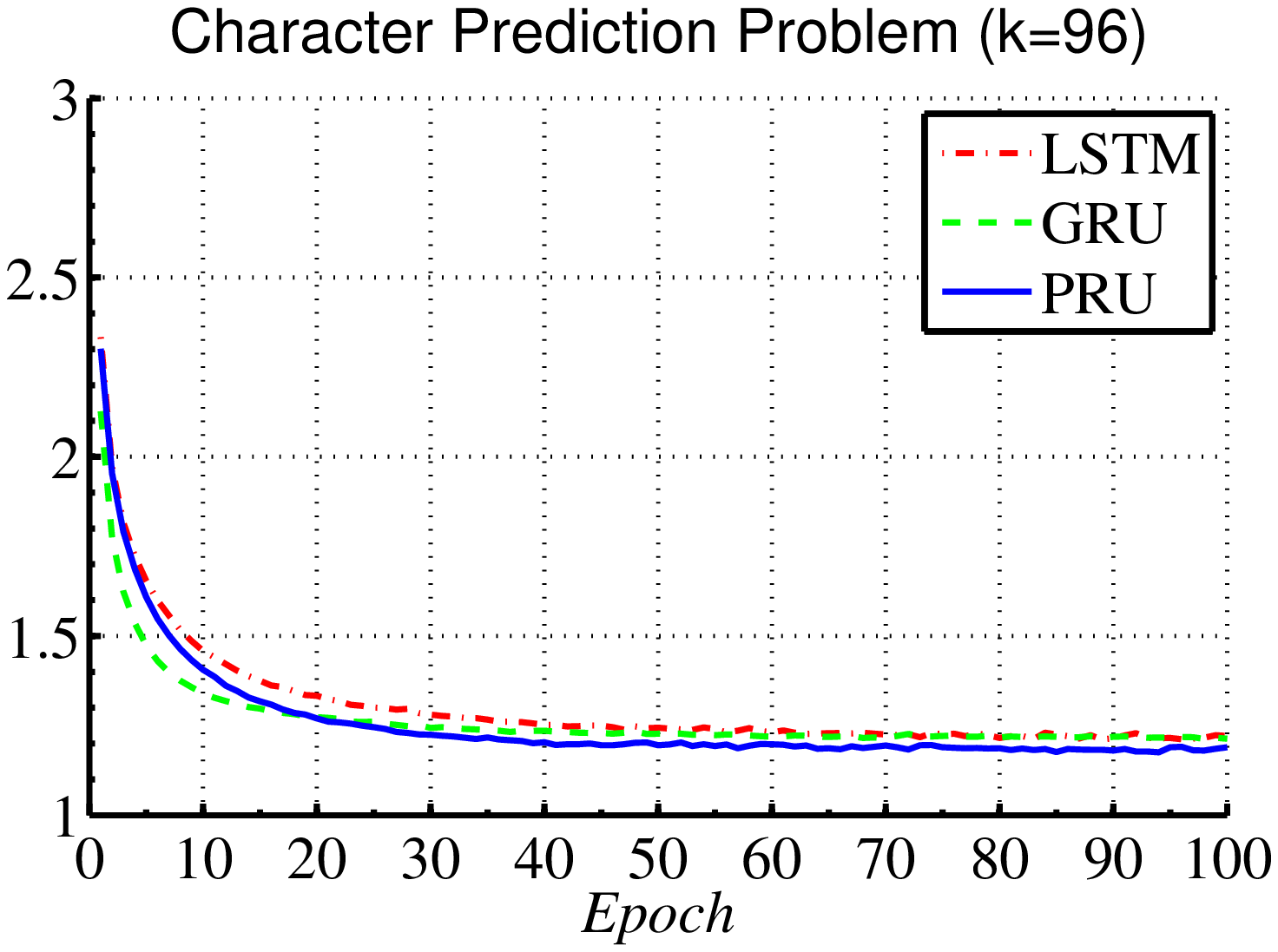} &
\includegraphics[width=.3\textwidth]{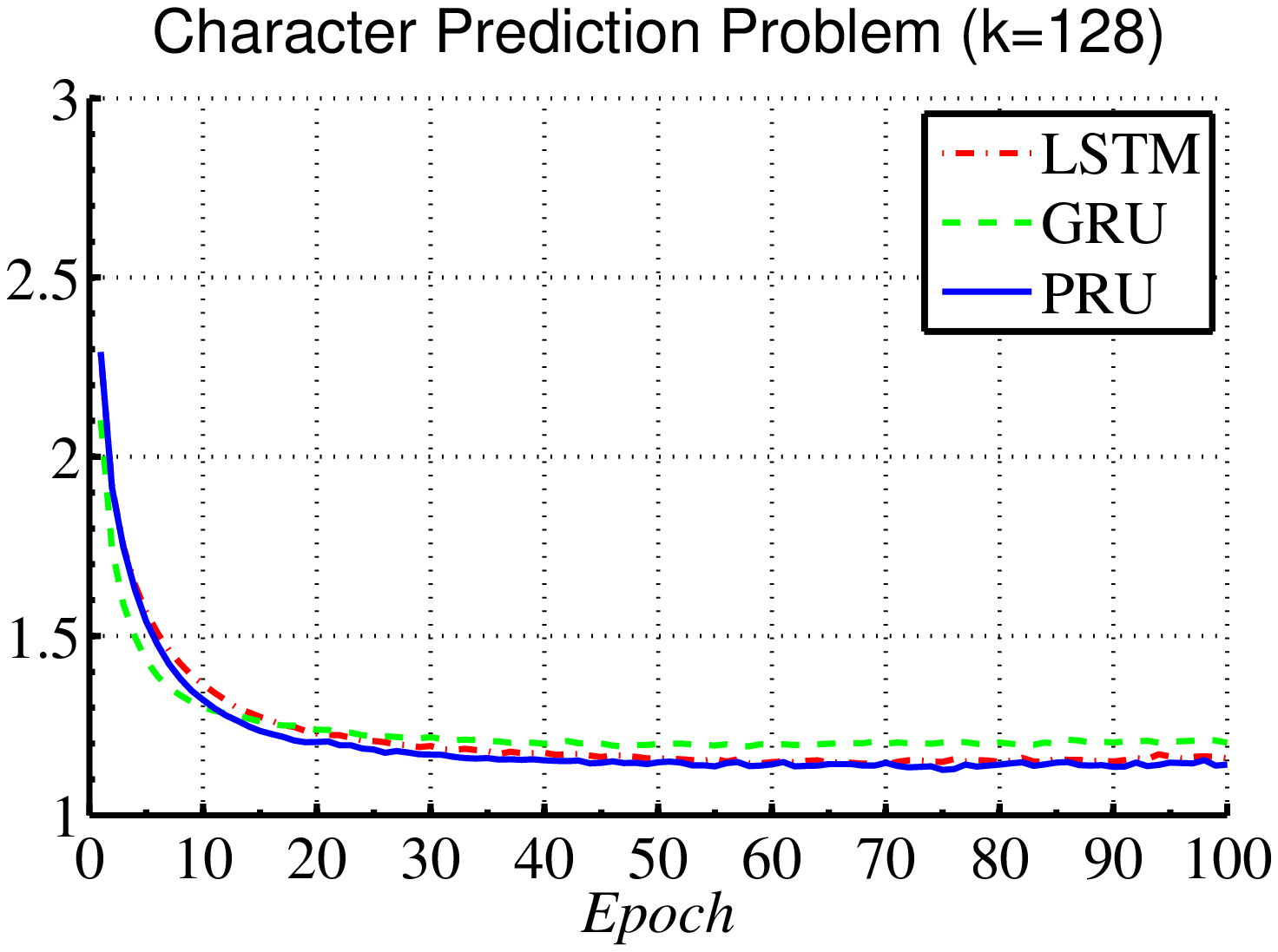}
\end{tabular}
\vspace{-0.3cm}
\caption{\label{fig:overall_CPP} CEL comparison of LSTM, GRU and PRU in Character Prediction Problem}
\end{figure}

\noindent \underline{\bf Results:}
%Results are obtained for LTSM, GRU and PRU under various settings of model state-space dimensions $k$.
Figure \ref{fig:overall_CPP} plots the performances of the three units as functions of SDG epoch number. The three units show very close performances for the chosen three settings of state space dimension. The table below lists the CEL performance of the three recurrent units at the end of SDG iterations, where PRU appears slightly outperform the other two.
%\begin{table}[!hbp]
\centerline{
\begin{tabular}{|@{\hspace{0.3cm}}c@{\hspace{0.3cm}\centering}|@{\hspace{0.3cm}}c|@{\hspace{0.3cm}}c|@{\hspace{0.3cm}}c|}
\hline
 State Space Dimension & LSTM & GRU & PRU \\
\hline
64 & 1.2752 & 1.3015 & 1.2245 \\
\hline
96 & 1.2211 & 1.2132 & 1.1894 \\
\hline
128 & 1.1584 & 1.1968 & 1.1410 \\
\hline
\end{tabular}
}
%\caption{\label{tab:cel}Cross-entropy loss for Character Prediction}
%\end{table}
The average training time for PRU, GRU, and LSTM per epoch are respectively $83.65$, $104.65$ and  $188.86$ seconds
respectively, with PRU leading by a significant margin.

\subsection{MNIST Image Classification Problem}

The MNIST dateset contains images of handwritten digits$('0'-'9')$, all the images are of size $28 \times 28$, we treat each row of the image(28 pixels) as a single
input in the input sequence. Let ${\cal M}_{\rm image}$ be a ``image-classification machine'', which takes an input sequence $(x_1, x_2, \ldots, x_N)$ with $N=28$ and predict the label $y$ of the input sequence. the machine is asked to predict the category of the image after seeing all the pixels and ${\cal M}_{\rm image}$ can be defined by
\[
{\cal M}_{\rm image}(x_1, \ldots, x_N):= y
\]

The objective of the MNIST Image Classfication Problem is then to train a model that simulates the behaviour of ${\cal M}_{\rm char}$.

\noindent \underline{\bf Modelling:} It is natural to take input space ${\pazocal X} = {\mathbb R}^{28}$ and ${\pazocal Y} = {\mathbb R}^{10}$, since the number of image category is $10$.
 in the models. The output $y$ is computed by a soft-max classifier.

\noindent \underline{\bf Dataset:} The MNIST dataset\footnote{ http://yann.lecun.com/exdb/mnist/} contains 60,000 images in the training set, and 10,000 in the test set.

\noindent \underline{\bf Training and Evaluation:}  The objective of this problem is to maximize the prediction accuraty (PA)
\[
{\cal E}_{\rm PA}(\theta)=
\frac{1}{K}
\sum_{i=1}^{K} \mathcal{G}({\cal M}_{\rm image} (x_1^i, \ldots, x_N^i), y_i)
\]
where we have used $K$ to denote the total number of examples , $(x_1^i, \ldots, x_N^i)$ and $y_i$ denote the input and label of un example respectively, if ${\cal M}_{\rm image} (x_1^i, \ldots, x_N^i)$ matches $y_i$, the value of $\mathcal{G}({\cal M}_{\rm image} (x_1^i, \ldots, x_N^i), y_i)$ is $1$, otherwise $0$.
 %$y_t$ is the output at the time step $t$ and $p_t$ is the target output.
 Mini-batched SGD with Adadelta dynamic learning rate \cite{zeiler2012adadelta} is used for optimization. All parameters are randomly initialized in $[-0.1, 0.1]$, and the base learning rate is set to $0.1$. PA is used to evaluate the models.

\begin{figure*}[htb!]\centering
\begin{tabular}{cc}
\includegraphics[width=.30\textwidth]{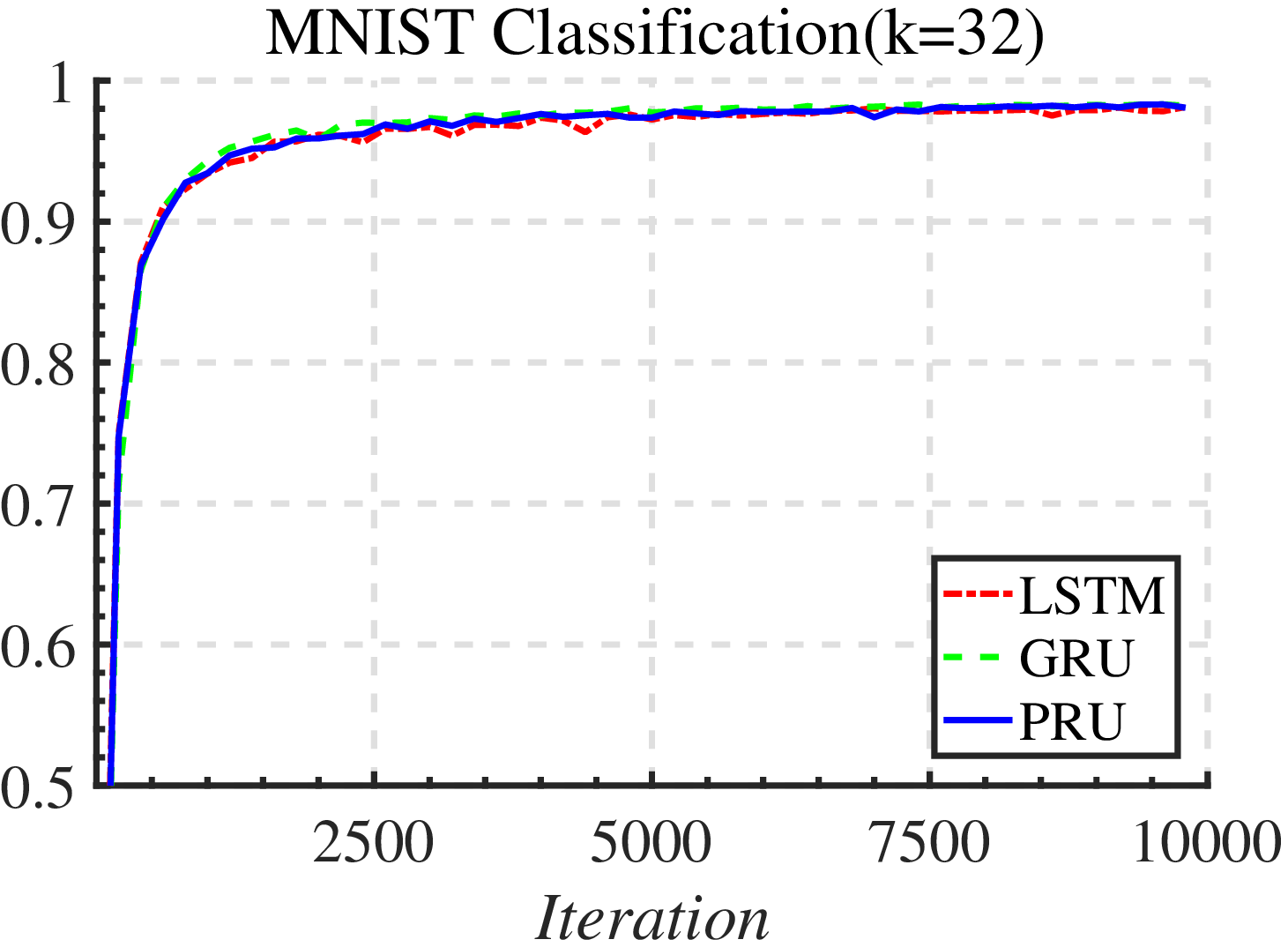} &
\includegraphics[width=.30\textwidth]{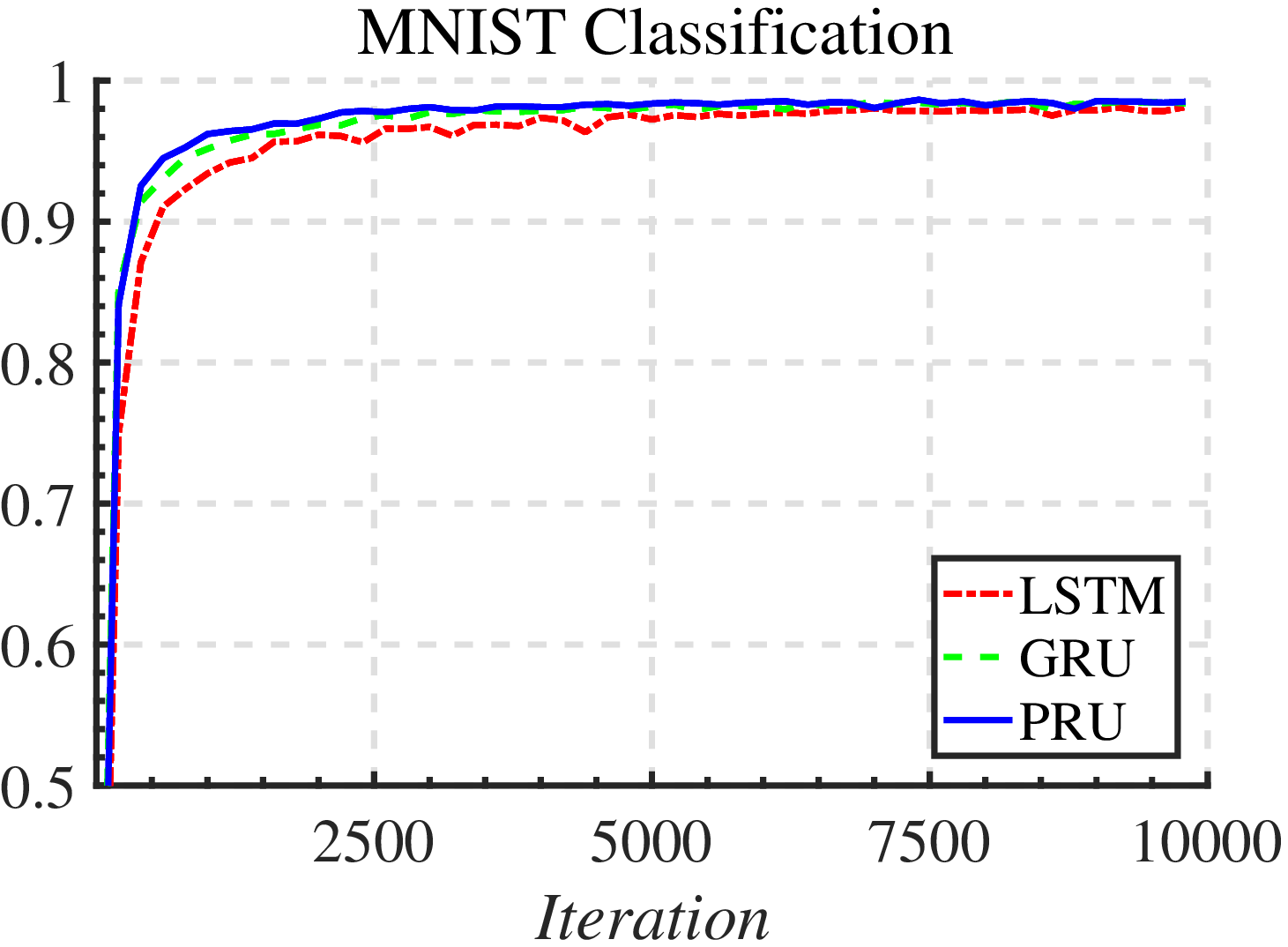}
\end{tabular}
\caption{\label{fig:overall_mnist} Test prediction accuracy comparison of LSTM, GRU and PRU in MNIST Image Classification Problem: Under same state space dimension(left), With same number of parameters(right)}
\end{figure*}

\begin{table}[htb!]
\centerline{
\begin{tabular}{|@{\hspace{0.3cm}}c@{\hspace{0.3cm}\centering}|@{\hspace{0.3cm}}c|@{\hspace{0.3cm}}c|@{\hspace{0.3cm}}c|}
\hline
 Experiment Setting & LSTM & GRU & PRU \\
\hline
Same state space dimension & 0.9815 & 0.9821 & 0.9852 \\
\hline
Same number of parameters & 0.9815 & 0.9843 & 0.9856 \\
\hline
\end{tabular}
}
\caption{\label{tab:cel}Prediction Accuracy for MNIST Image Classification}
\end{table}
\noindent \underline{\bf Results:}
Results are obtained for LTSM, GRU and PRU in two conditions: under of model state-space dimensions $k$ and the same number of parameters $p$.
Figure \ref{fig:overall_mnist} plots the performances of the three units as functions of iteration number. The three units show very close performances for the chosen three settings of state space dimension. The table below lists the PA performance of the three recurrent units at the end of SDG iterations, the performance on MNIST is similar to that on the character prediction problem, where PRU appears slightly outperform the other two.

The average training time for PRU, GRU, and LSTM per epoch are respectively $110.73$, $154.94$ and  $210.04$ seconds
respectively, with PRU leading by a significant margin.

%\vspace{-0.2cm}

\section{Concluding Remarks}

This paper presents a new recurrent unit, PRU. Having very simple structure, PRU is shown to perform similarly to LSTM and GRU. This potentially allows  the use of PRU as a prototypical example for analytic study of LSTM-like recurrent networks. Its  complexity advantage may also make it a practical alternative to LSTM and GRU. 

This work is only the beginning of a journey towards understanding recurrent networks. It is our hope that PRU may provide some convenience to this important endeavor.

\clearpage

\bibliographystyle{plain}

\bibliography{DeepLearning}

\begin{thebibliography}{10}

\bibitem{bahdanau2014neural}
Dzmitry Bahdanau, Kyunghyun Cho, and Yoshua Bengio.
\newblock Neural machine translation by jointly learning to align and
  translate.
\newblock {\em Computer Science}, 2014.

\bibitem{bengio1994learning}
Yoshua Bengio, Patrice Simard, and Paolo Frasconi.
\newblock Learning long-term dependencies with gradient descent is difficult.
\newblock {\em IEEE transactions on neural networks}, 5(2):157--166, 1994.

\bibitem{cheng2016long}
Jianpeng Cheng, Li~Dong, and Mirella Lapata.
\newblock Long short-term memory-networks for machine reading.
\newblock {\em arXiv preprint arXiv:1601.06733}, 2016.

\bibitem{cho2014learning}
Kyunghyun Cho, Bart Van~Merri{\"e}nboer, Caglar Gulcehre, Dzmitry Bahdanau,
  Fethi Bougares, Holger Schwenk, and Yoshua Bengio.
\newblock Learning phrase representations using rnn encoder-decoder for
  statistical machine translation.
\newblock pages 1724--1735, 2014.

\bibitem{chung2014empirical}
Junyoung Chung, Caglar Gulcehre, KyungHyun Cho, and Yoshua Bengio.
\newblock Empirical evaluation of gated recurrent neural networks on sequence
  modeling.
\newblock {\em arXiv preprint arXiv:1412.3555}, 2014.

\bibitem{elman1991distributed}
Jeffrey~L Elman.
\newblock Distributed representations, simple recurrent networks, and
  grammatical structure.
\newblock {\em Machine learning}, 7(2-3):195--225, 1991.

\bibitem{gers2001long}
Felix Gers.
\newblock {\em Long short-term memory in recurrent neural networks}.
\newblock PhD thesis, Universit{\"a}t Hannover, 2001.

\bibitem{gers2000learning}
Felix~A Gers, J{\"u}rgen Schmidhuber, and Fred Cummins.
\newblock Learning to forget: Continual prediction with lstm.
\newblock {\em Neural computation}, 12(10):2451--2471, 2000.

\bibitem{peephole2002learning}
Felix~A Gers, Nicol~N Schraudolph, and J{\"u}rgen Schmidhuber.
\newblock Learning precise timing with lstm recurrent networks.
\newblock {\em Journal of machine learning research}, 3(Aug):115--143, 2002.

\bibitem{graves2013speech}
Alex Graves, Abdel-rahman Mohamed, and Geoffrey Hinton.
\newblock Speech recognition with deep recurrent neural networks.
\newblock In {\em 2013 IEEE international conference on acoustics, speech and
  signal processing}, pages 6645--6649. IEEE, 2013.

\bibitem{greff2015lstm}
K~Greff, R.~K. Srivastava, J~Koutnik, B.~R. Steunebrink, and J~Schmidhuber.
\newblock Lstm: A search space odyssey.
\newblock {\em IEEE Transactions on Neural Networks and Learning Systems},
  2015.

\bibitem{HintonSalakhutdinov_Science_2006:deepBeliefNet}
G.~E. Hinton and R.~R. Salakhutdinov.
\newblock Reducing the dimensionality of data with neural networks.
\newblock {\em Science}, 313(5786):504--507, 2006.

\bibitem{hochreiter1997long}
Sepp Hochreiter and J{\"u}rgen Schmidhuber.
\newblock Long short-term memory.
\newblock {\em Neural computation}, 9(8):1735--1780, 1997.

\bibitem{jozefowicz2015empirical}
Rafal Jozefowicz, Wojciech Zaremba, and Ilya Sutskever.
\newblock An empirical exploration of recurrent network architectures.
\newblock In {\em Proceedings of The 32nd International Conference on Machine
  Learning}, pages 2342--2350, 2015.

\bibitem{Kha02}
H.~K. Khalil.
\newblock {\em Nonlinear Systems}.
\newblock Prentice-Hall, Englewood Cliffs, NJ, 3nd edition, 2002.

\bibitem{kim2015character}
Yoon Kim, Yacine Jernite, David Sontag, and Alexander~M Rush.
\newblock Character-aware neural language models.
\newblock 2016.

\bibitem{lecun2015deepNatureReview}
Yann LeCun, Yoshua Bengio, and Geoffrey Hinton.
\newblock Deep learning.
\newblock {\em Nature}, 521(7553):436--444, 2015.

\bibitem{mikolov2014learning}
Tomas Mikolov, Armand Joulin, Sumit Chopra, Michael Mathieu, and Marc'Aurelio
  Ranzato.
\newblock Learning longer memory in recurrent neural networks.
\newblock {\em arXiv preprint arXiv:1412.7753}, 2014.

\bibitem{mikolov2010recurrent}
Tomas Mikolov, Martin Karafi{\'a}t, Lukas Burget, Jan Cernock{\`y}, and Sanjeev
  Khudanpur.
\newblock Recurrent neural network based language model.
\newblock In {\em Interspeech}, volume~2, page~3, 2010.

\bibitem{pascanu2013difficulty}
Razvan Pascanu, Tomas Mikolov, and Yoshua Bengio.
\newblock On the difficulty of training recurrent neural networks.
\newblock {\em ICML (3)}, 28:1310--1318, 2013.

\bibitem{serban2016building}
Iulian~V Serban, Alessandro Sordoni, Yoshua Bengio, Aaron Courville, and Joelle
  Pineau.
\newblock Building end-to-end dialogue systems using generative hierarchical
  neural network models.
\newblock In {\em Proceedings of the 30th AAAI Conference on Artificial
  Intelligence (AAAI-16)}, 2016.

\bibitem{srivastava2015unsupervised}
Nitish Srivastava, Elman Mansimov, and Ruslan Salakhutdinov.
\newblock Unsupervised learning of video representations using lstms.
\newblock {\em CoRR, abs/1502.04681}, 2, 2015.

\bibitem{van2016pixel}
Aaron van~den Oord, Nal Kalchbrenner, and Koray Kavukcuoglu.
\newblock Pixel recurrent neural networks.
\newblock {\em arXiv preprint arXiv:1601.06759}, 2016.

\bibitem{zeiler2012adadelta}
Matthew~D Zeiler.
\newblock Adadelta: an adaptive learning rate method.
\newblock {\em arXiv preprint arXiv:1212.5701}, 2012.

\end{thebibliography}

\end{document}